\RequirePackage{fix-cm}
\documentclass[twocolumn, natbib]{svjour3}          
\smartqed  
\usepackage{graphicx}
\usepackage{mathptmx}      
\usepackage{subfigure}
\usepackage{pifont}
\newcommand{\cmark}{\ding{52}}%
\newcommand{\xmark}{\ding{56}}%
\usepackage{amssymb} 
\usepackage{tabularx}
\usepackage{multirow}
\usepackage{color}
\definecolor{green}{rgb}{0.0, 0.5, 0.0}
\definecolor{orange}{rgb}{1, 0.5, 0}
\usepackage[colorlinks]{hyperref} 
\hypersetup{
allcolors={blue}
}
\usepackage{etoolbox}

\makeatletter

\patchcmd{\NAT@citex}
  {\@citea\NAT@hyper@{%
     \NAT@nmfmt{\NAT@nm}%
     \hyper@natlinkbreak{\NAT@aysep\NAT@spacechar}{\@citeb\@extra@b@citeb}%
     \NAT@date}}
  {\@citea\NAT@nmfmt{\NAT@nm}%
   \NAT@aysep\NAT@spacechar\NAT@hyper@{\NAT@date}}{}{}

\patchcmd{\NAT@citex}
  {\@citea\NAT@hyper@{%
     \NAT@nmfmt{\NAT@nm}%
     \hyper@natlinkbreak{\NAT@spacechar\NAT@@open\if*#1*\else#1\NAT@spacechar\fi}%
       {\@citeb\@extra@b@citeb}%
     \NAT@date}}
  {\@citea\NAT@nmfmt{\NAT@nm}%
   \NAT@spacechar\NAT@@open\if*#1*\else#1\NAT@spacechar\fi\NAT@hyper@{\NAT@date}}
  {}{}

\makeatother

\begin{document}
\sloppy

\title{Pixel-in-Pixel Net: Towards Efficient Facial Landmark Detection in the Wild
}


\author{Haibo Jin$^1$         \and
        Shengcai Liao$^1$\thanks{Shengcai Liao is the corresponding author}     \and
        Ling Shao$^{1,2}$ 
}


\institute{
           Haibo Jin \at
           \email{haibo.nick.jin@gmail.com}  
           \and
           Shengcai Liao \at
           \email{scliao@ieee.org} 
           \and
           Ling Shao \at
           \email{ling.shao@ieee.org}      
           \and           
           \at $^1$ Inception Institute of Artificial Intelligence (IIAI), \\
           Abu Dhabi, UAE  
           \and
           \at $^2$ Mohamed bin Zayed University \\
           of Artificial Intelligence (MBZUAI), \\
           Abu Dhabi, UAE  
}

\date{Received: date / Accepted: date}

\maketitle

\begin{abstract}

Recently, heatmap regression models have become popular due to their superior performance in locating facial landmarks. However, three major problems still exist among these models: (1) they are computationally expensive; (2) they usually lack explicit constraints on global shapes; (3) domain gaps are commonly present. To address these problems, we propose Pixel-in-Pixel Net (PIPNet) for facial landmark detection. The proposed model is equipped with a novel detection head based on heatmap regression, which conducts score and offset predictions simultaneously on low-resolution feature maps. By doing so, repeated upsampling layers are no longer necessary, enabling the inference time to be largely reduced without sacrificing model accuracy. Besides, a simple but effective neighbor regression module is proposed to enforce local constraints by fusing predictions from neighboring landmarks, which enhances the robustness of the new detection head. To further improve the cross-domain  generalization capability of PIPNet, we propose self-training with curriculum. This  training strategy is able to mine more reliable pseudo-labels from unlabeled data across domains by starting with an easier task, then gradually increasing the difficulty to provide more precise labels. Extensive experiments demonstrate the superiority of PIPNet, which obtains state-of-the-art results on three out of six popular benchmarks under the supervised setting. The results on two cross-domain test sets are also consistently improved compared to the baselines. Notably, our lightweight version of PIPNet runs at 35.7 FPS and 200 FPS on CPU and GPU, respectively, while still maintaining a competitive accuracy to state-of-the-art methods. The code of PIPNet is available at \href{https://github.com/jhb86253817/PIPNet}{https://github.com/jhb86253817/PIPNet}.  
\keywords{Facial landmark detection \and Pixel-in-pixel regression \and Self-training with curriculum \and Unsupervised domain adaptation}
\end{abstract}

\section{Introduction}
\label{sec:1}

\begin{figure*}
\centering
    \subfigure[CPU\label{fig:speed_cpu}]{
    \includegraphics[width=0.48\linewidth]{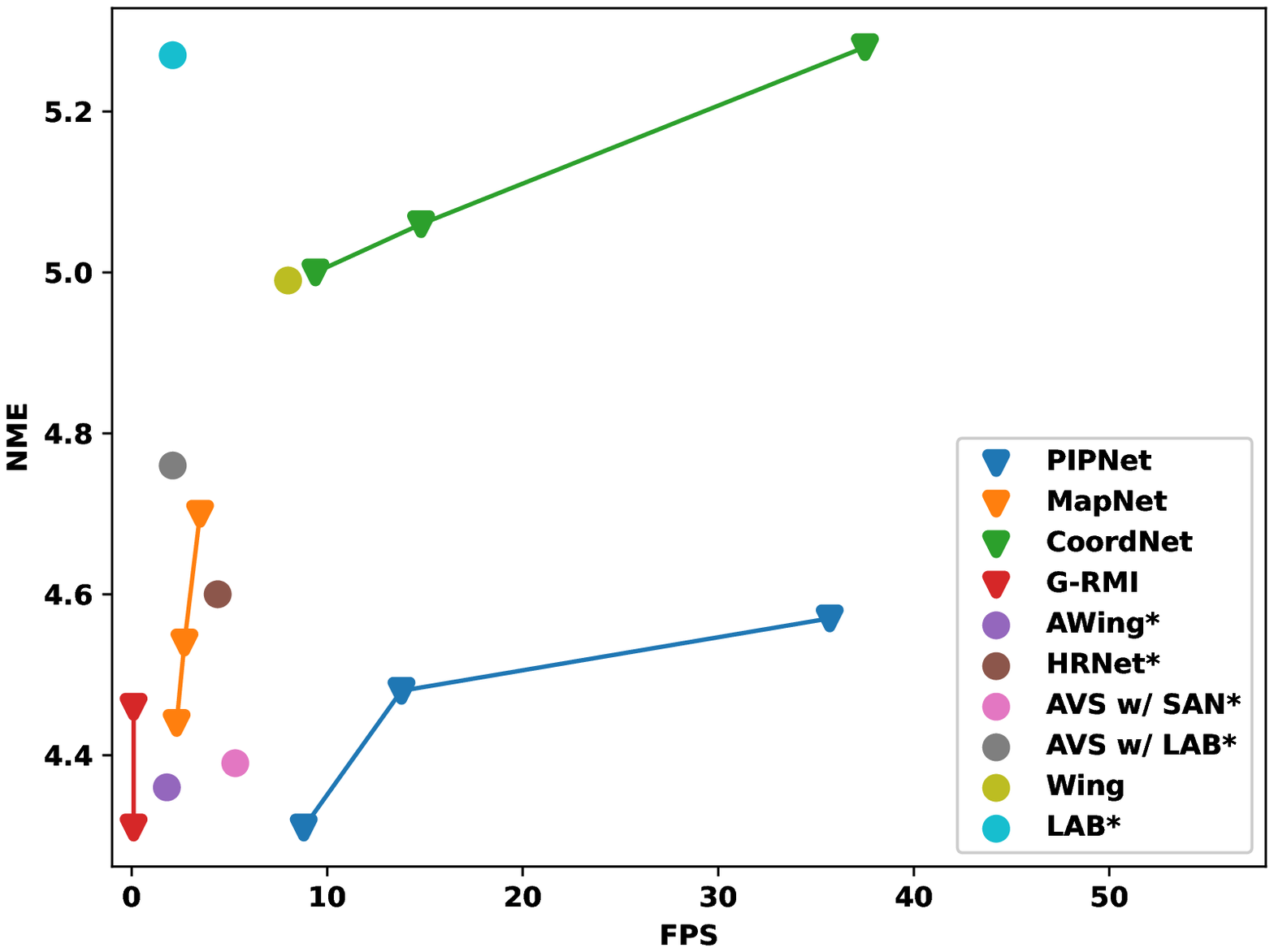}
    }         
    \subfigure[GPU\label{fig:speed_gpu}]{
    \includegraphics[width=0.48\linewidth]{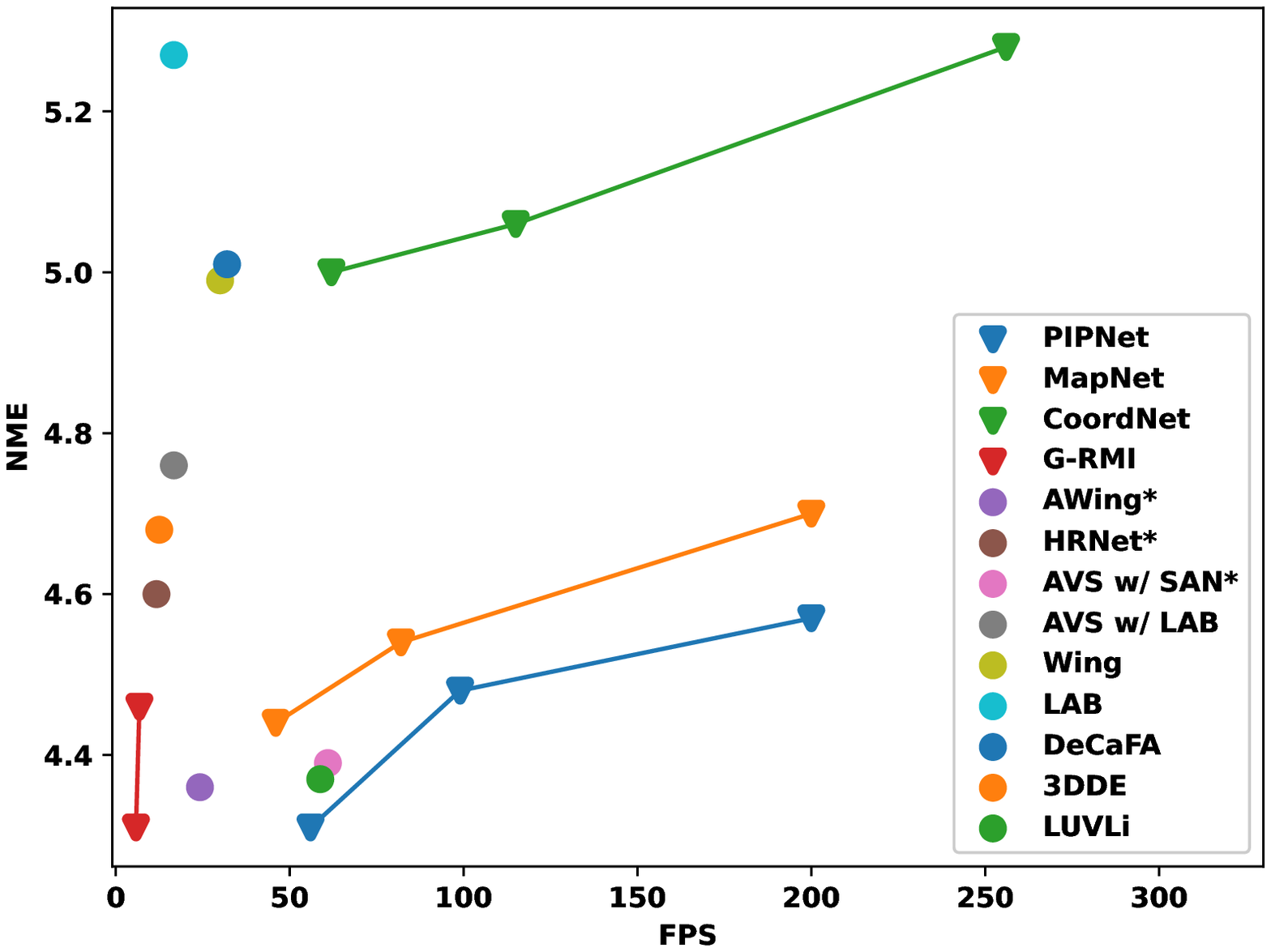}
    }        
    \caption{Comparison with the existing methods in terms of speed-accuracy trade-off. The NMEs (\%) are tested on the WFLW test set. The closer a model is to the bottom-right corner, the better its speed-accuracy trade-off. The existing methods with * were tested by us under the same environment as our methods. (a) Tested on CPU. (b) Tested on GPU. \label{fig:speed}}
\end{figure*}

Facial landmark detection aims to locate predefined landmarks on a human face, the results of which are useful for several face analysis tasks, such as face recognition~\citep{TYR14, LWY17, LJL13}, face tracking~\citep{KMT17}, face editing~\citep{TZS16}, etc. These applications usually run on online systems in uncontrolled environments, requiring facial landmark detectors to be accurate, robust, and computationally efficient, all at the same time.

Over the last few years, significant progress has been made in this area, especially by deep convolutional neural networks (CNNs) that can be trained end-to-end. Among recent works, some~\citep{FKA18, WBF19} aim at improving loss functions, some~\citep{DYO18, QSW19} focus on data augmentation for better generalization, and others~\citep{WQY18, LZH19} address the semantic ambiguity issue. However, few studies focusing on detection heads have been conducted, despite their essentialness to landmark detectors. Specifically, the detection head can affect the accuracy, robustness, and efficiency of a model. For deep learning based facial landmark detection, there are two widely used detection heads, namely heatmap regression and coordinate regression. Heatmap regression can achieve good results, but it has two drawbacks: (1) it is computationally expensive; (2) it is sensitive to outliers (see Figure~\ref{fig:WFLW_vis_map}). In contrast, coordinate regression is fast and robust, but not accurate enough (see Figure~\ref{fig:WFLW_vis_coord}). Although coordinate regression can be used in a multi-stage manner to yield better performance, its inference speed becomes slow as a result. Accordingly, in this work, we aim to answer the following question: Is there a detection head that possesses the advantages of both heatmap regression and coordinate regression?

\begin{figure}
\centering
    \subfigure[GSL\label{fig:paradigm_1}]{
    \includegraphics[width=0.4\linewidth]{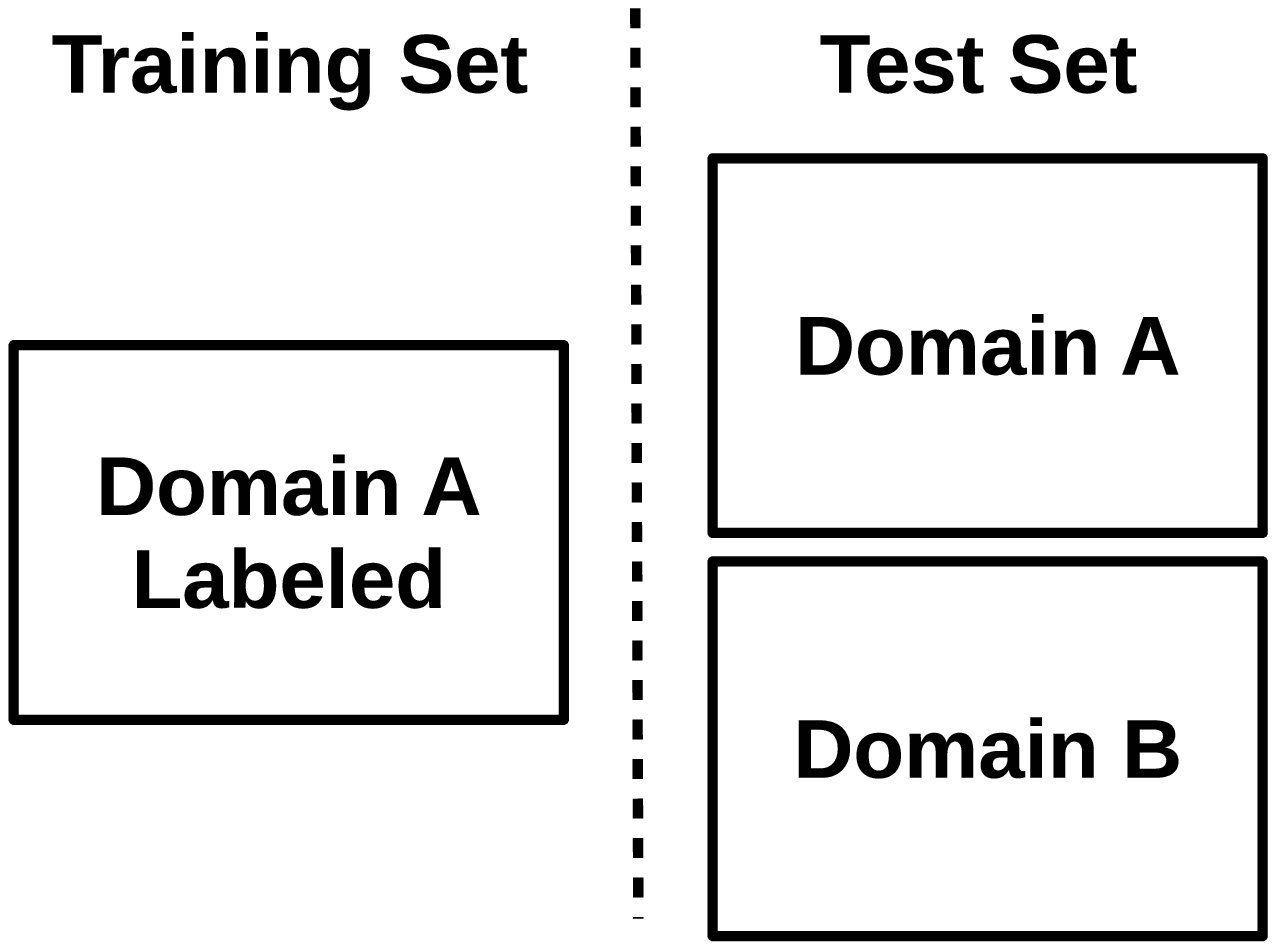}
    } 
    \subfigure[SSL\label{fig:paradigm_2}]{
    \includegraphics[width=0.4\linewidth]{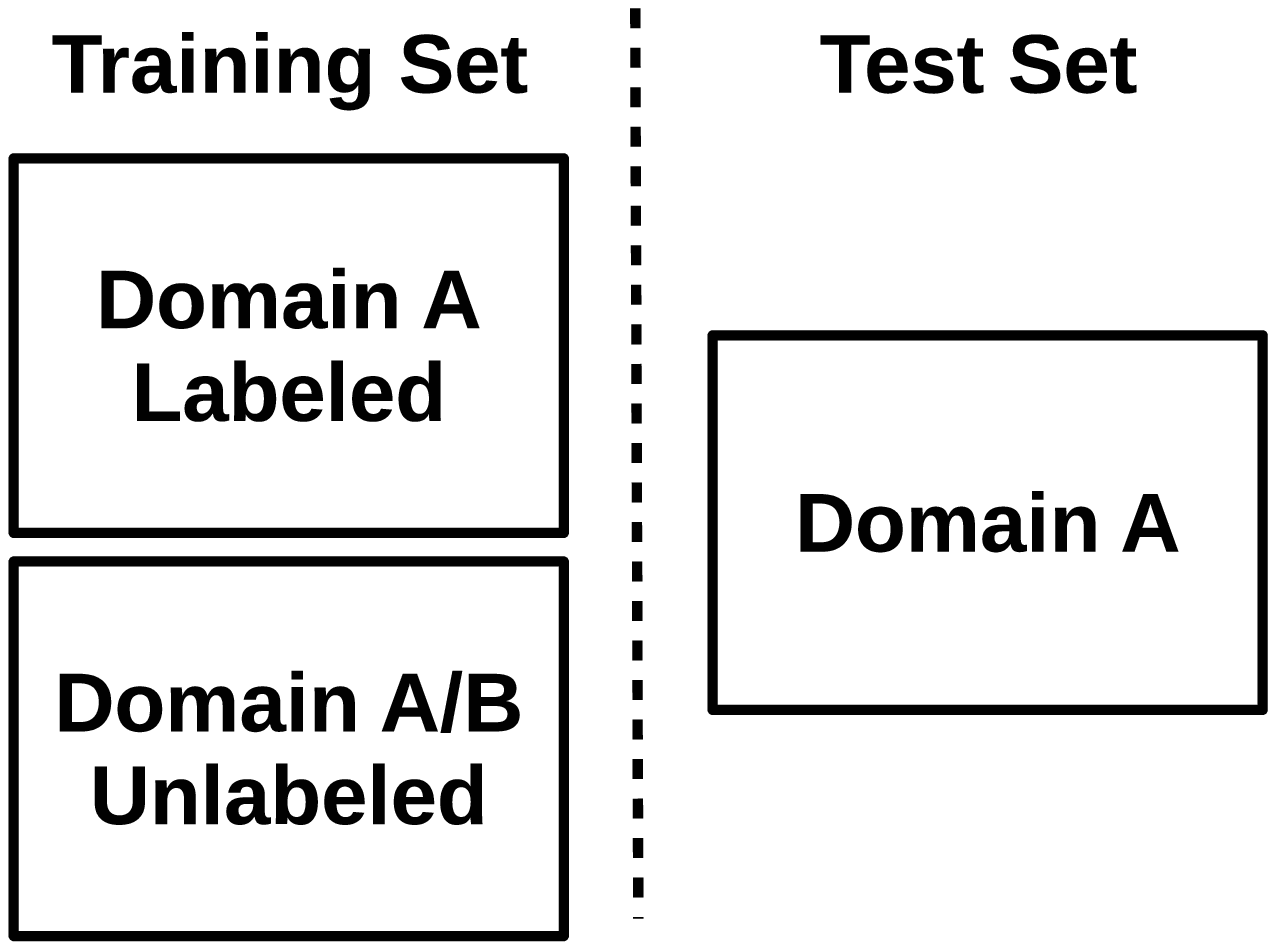}
    }

    \subfigure[UDA\label{fig:paradigm_3}]{
    \includegraphics[width=0.4\linewidth]{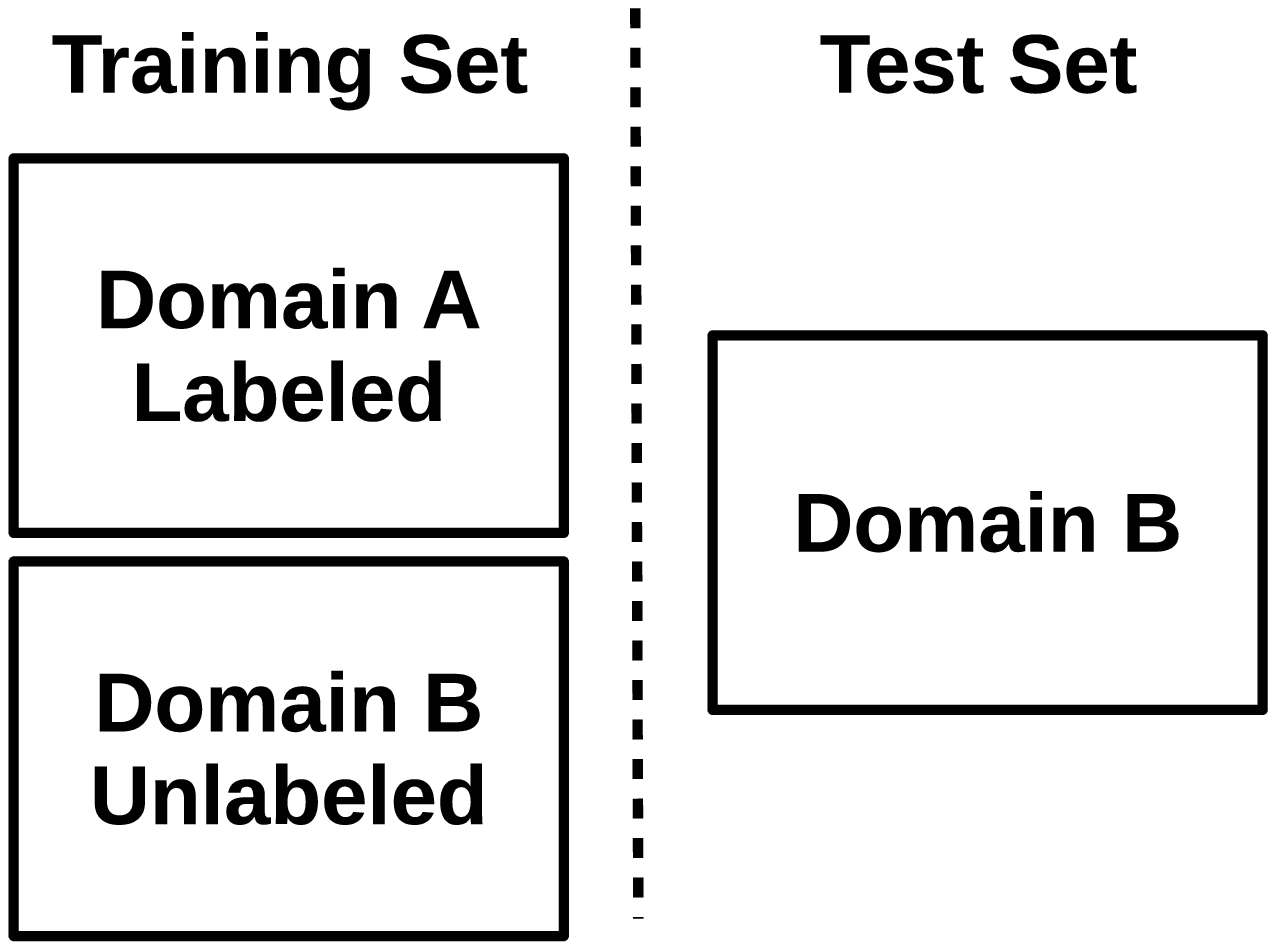}
    }
    \subfigure[GSSL\label{fig:paradigm_4}]{
    \includegraphics[width=0.4\linewidth]{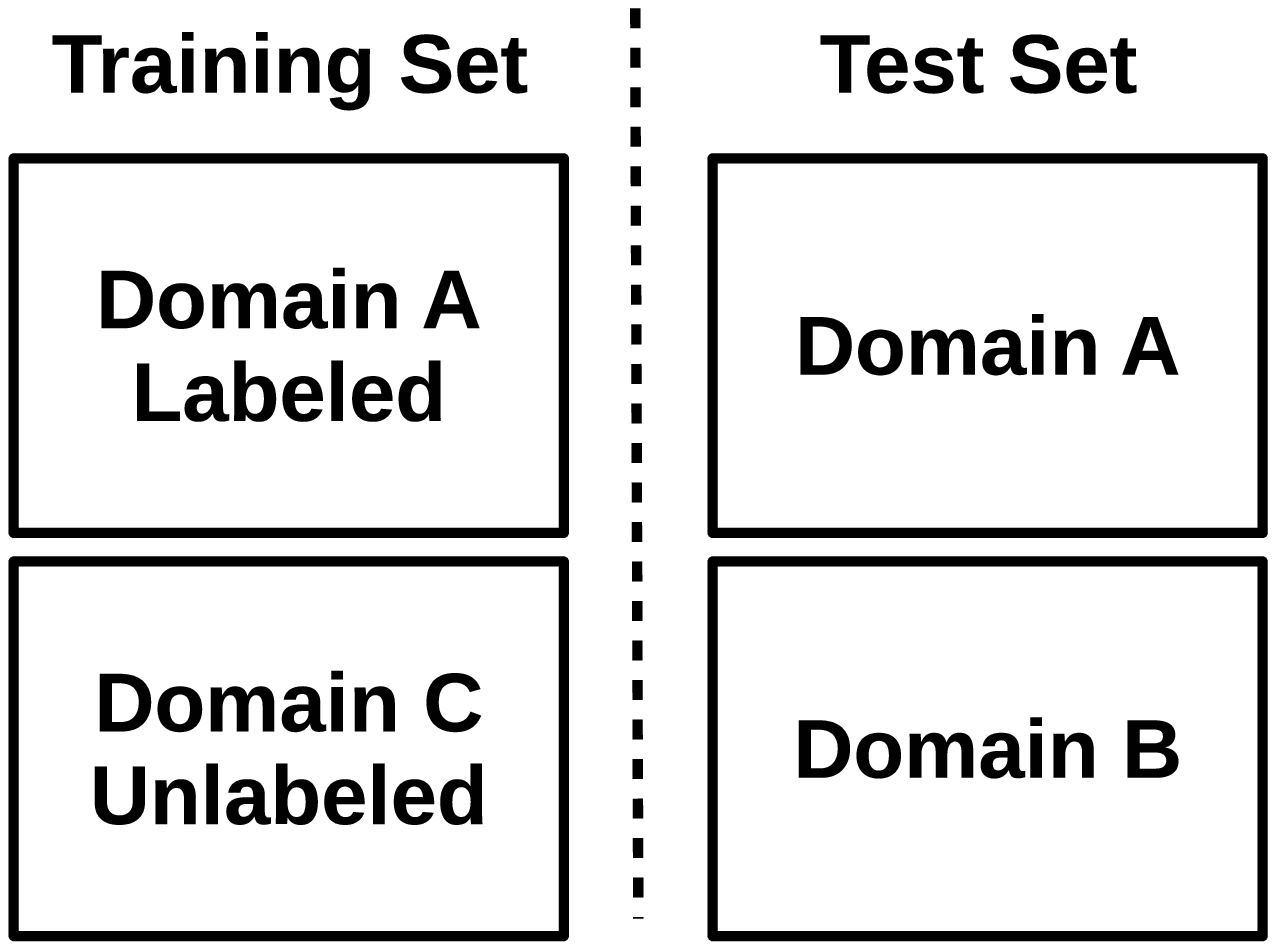}
    }
    \caption{Different training and testing paradigms. (a) Generalizable supervised learning. (b) Semi-supervised learning. (c) Unsupervised domain adaptation. (d) Generalizable semi-supervised learning.}
    \label{fig:paradigm}
\end{figure}

Generalization capability across domains is another challenge of facial landmark detection. As shown in~\citep{VBV19}, there are great performance gaps between intra-domain and cross-domain test sets. For a model to perform robustly under unconstrained environments, the domain gaps should be made as small as possible. Existing works~\citep{WQY18,ZSZ19,QSW19} all address this problem by training a model with supervised learning, and then directly evaluating it on cross-domain datasets. We call this paradigm generalizable supervised learning (GSL). A drawback of GSL is that it relies on human-designed modules for cross-domain generalization, which are not scalable. One may suggest training the models on various datasets with supervised learning, but this is impractical due to the high labor costs of annotation. Therefore, we have decided to explore generalizable semi-supervised learning (GSSL) for facial landmark detection, which utilizes both labeled and unlabeled data across domains to obtain better generalization capability. Compared to GSL, GSSL is more scalable because it is data-driven, and unlabeled images are relatively easy to collect. Unsupervised domain adaptation (UDA), a special case of GSSL, has been successfully adopted in several vision tasks, including image classification~\citep{LCW15,GaL15,KJY19}, object detection~\citep{CLS18,ZPY19,SUH19}, person re-identification~\citep{PXW16,YWZ17,ZZL18,DZY18,YZW19,ZLX20}, and so on. However, the effectiveness of UDA for facial landmark detection remains unknown. Figure~\ref{fig:paradigm} shows the difference between various training and testing paradigms. As shown in the figure, the main difference between GSSL and UDA is that GSSL is not very strict about the domain of the unlabeled data, while UDA usually requires the unlabeled and test data to be from the same domain. In this work, we investigate the feasibility of GSSL (including UDA) for better cross-domain generalization on facial landmark detection. 

In order to obtain an efficient facial landmark detector that can run in the wild, we propose a new model named Pixel-In-Pixel Net (PIPNet). PIPNet consists of three essential parts: (1) Pixel-In-Pixel (PIP) regression; (2) a neighbor regression module; and (3) self-training with curriculum. PIP regression, the detection head of PIPNet, is based on heatmap regression, but further predicts offsets within each feature map pixel (grid) in addition to predicting scores. By doing so, the model can still achieve good results even when the stride of the network is large (i.e., the last feature map is of low resolution). Consequently, the upsampling layers for heatmap regression can be eliminated to save considerable computational cost, without sacrificing accuracy. The neighbor regression module is designed to enhance the robustness of the PIP regression, inspired by coordinate regression (see Section~\ref{sec:3.2}). For each landmark, the neighbor regression module predicts the locations of the neighboring landmarks within each feature map pixel. The predicted neighbors are then merged with the results of PIP regression during inference. With marginal extra cost, the neighbor regression module is able to improve the robustness of the model by introducing local constraints on the shapes of the predicted landmarks.  With the help of PIP regression and the neighbor regression module, the proposed model inherits the advantages of both heatmap and coordinate regression. In fact, in Section~\ref{sec:3.1}, we show that heatmap and coordinate regression can be seen as two special cases of PIP regression with different strides. We also demonstrate the superiority of PIP regression (with neighbor regression) over the two alternatives in terms of bias-variance trade-off in Section~\ref{sec:4.3.2}. In order to better utilize unlabeled data across domains, we propose self-training with curriculum for generalizable semi-supervised learning. Different from standard self-training, self-training with curriculum starts with an easier task for the unlabeled data, and then gradually increases the difficulty to obtain more refined pseudo-labels. In this way, less errors are introduced from the estimated pseudo-labels, easing the mistake reinforcement problem of self-training. 

Our contributions in this work are summarized as follows.

\begin{enumerate}
\item We propose PIP regression as a novel detection head for facial landmark detection, which achieves comparable accuracy to heatmap regression, but runs significantly faster on CPU (see Figure~\ref{fig:speed_cpu}). We also show that PIP regression is a generalization of the two popular detection heads. To the best of our knowledge, this is the first study in this area that discusses the connection between heatmap and coordinate regression.
\item A neighbor regression module is proposed to enhance the robustness of the PIP regression, especially on cross-domain datasets. Additionally, we show that PIP regression with the neighbor regression module yields better performance than the two alternatives from the perspective of bias-variance trade-off.
\item Aiming to further improve the generalization capability of PIPNet on unseen domains, a new method is designed under the GSSL paradigm, termed self-training with curriculum. Experiments show that self-training with curriculum achieves consistently better results than its baselines on two cross-domain datasets. As far as we know, this is the first study to utilize unlabeled data to improve generalization capability on facial landmark detection.
\item We observe that CNN-based landmark detectors make predictions using not only semantic features, but also positional features. Specifically, even if the input image does not contain a human face, the landmark detectors still give face-like predictions (see Figure~\ref{fig:prior_2}-\ref{fig:prior_4}), which can be seen as an implicit prior learned from data. Coordinate regression has a stronger implicit prior than heatmap regression, which also explains the different characteristics of the two detection heads.  
\item The proposed PIPNet obtains state-of-the-art results on COFW, WFLW, and 300VW. Notably, PIPNet with ResNet-18 is able to run at 35.7 FPS and 200 FPS on CPU and GPU, respectively (see Figure\ref{fig:speed}), while its accuracy is still competitive with state-of-the-art methods.  
\end{enumerate}

The rest of the paper is organized as follows. Section~\ref{sec:2} briefly reviews the related works. Section~\ref{sec:3} introduces the proposed methods. The experimental results are presented in Section~\ref{sec:4}. Finally, we draw conclusions in Section~\ref{sec:5}.

\section{Related Work}
\label{sec:2}
In this section, we review relevant works on deeply supervised facial landmark detection (coordinate regression models and heatmap regression models), semi-supervised facial landmark detection, and the generalization capability across domains in this area.

\textbf{Coordinate Regression Models.} Coordinate regression can be used to directly map an input image to landmark coordinates. In the context of deep learning, the features of the input image are usually extracted using a CNN, then mapped to coordinates through fully connected layers. Due to its fixed connections to specific locations of feature maps, the end-to-end prediction of coordinate regression is inaccurate and biased. Therefore, coordinate regression is usually cascaded~\citep{SWT13,ZLL15,TSN16,LSX17,FKA18}, integrated with extra modules~\citep{WQY18,ZSZ19}, or built upon heatmap regression~\citep{VBV18, LZH19}. 

\begin{figure}
\centering
    \subfigure[Coordinate Regression\label{fig:heads_arc_1}]{
    \includegraphics[width=0.45\linewidth]{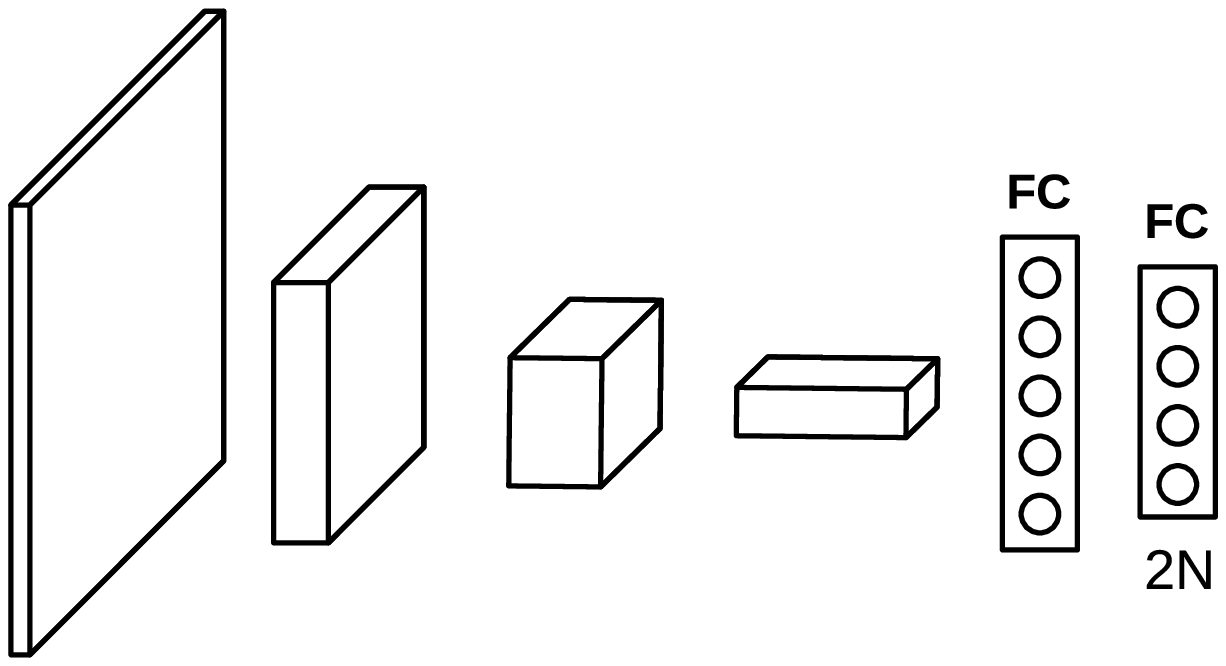}
    } 
    \subfigure[Heatmap Regression\label{fig:heads_arc_2}]{
    \includegraphics[width=0.45\linewidth]{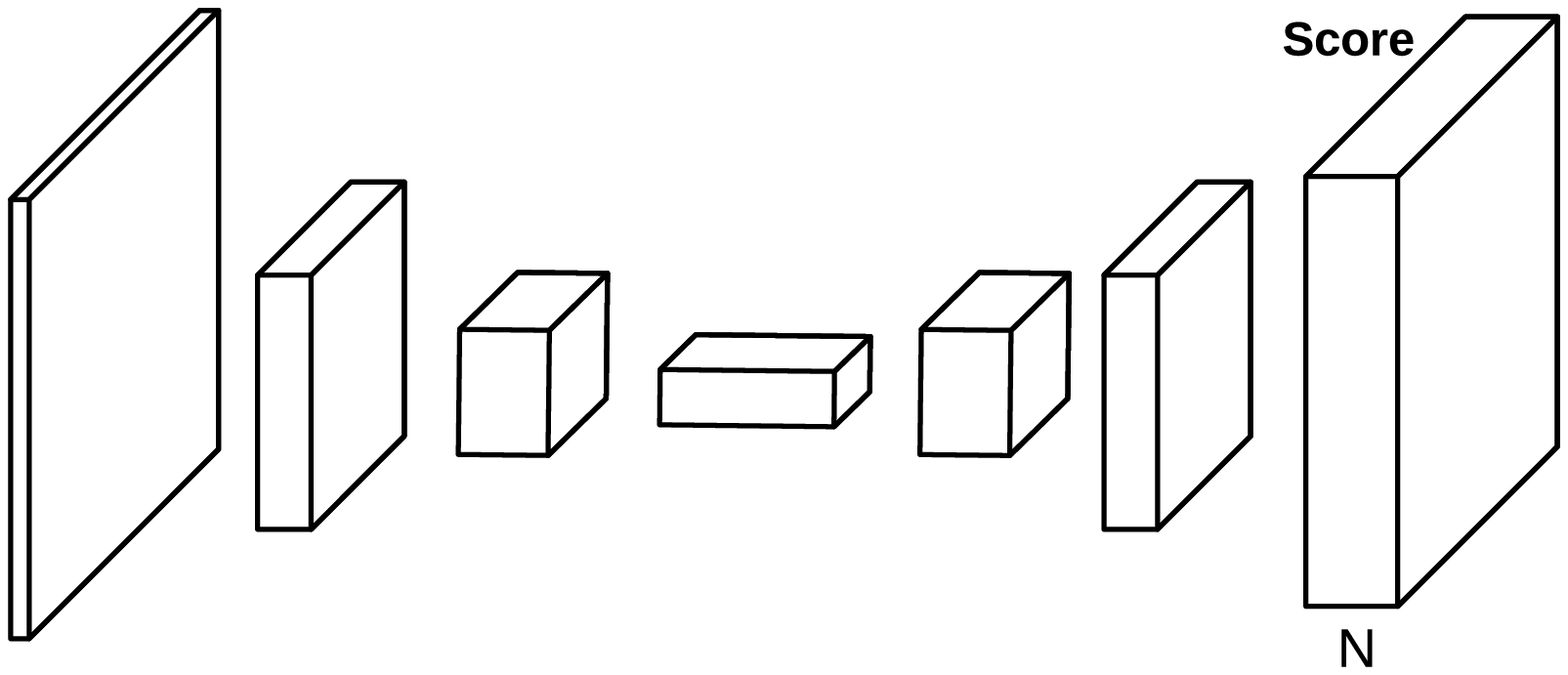}
    }

    \subfigure[PIP Regression\label{fig:heads_arc_3}]{
    \includegraphics[width=0.45\linewidth]{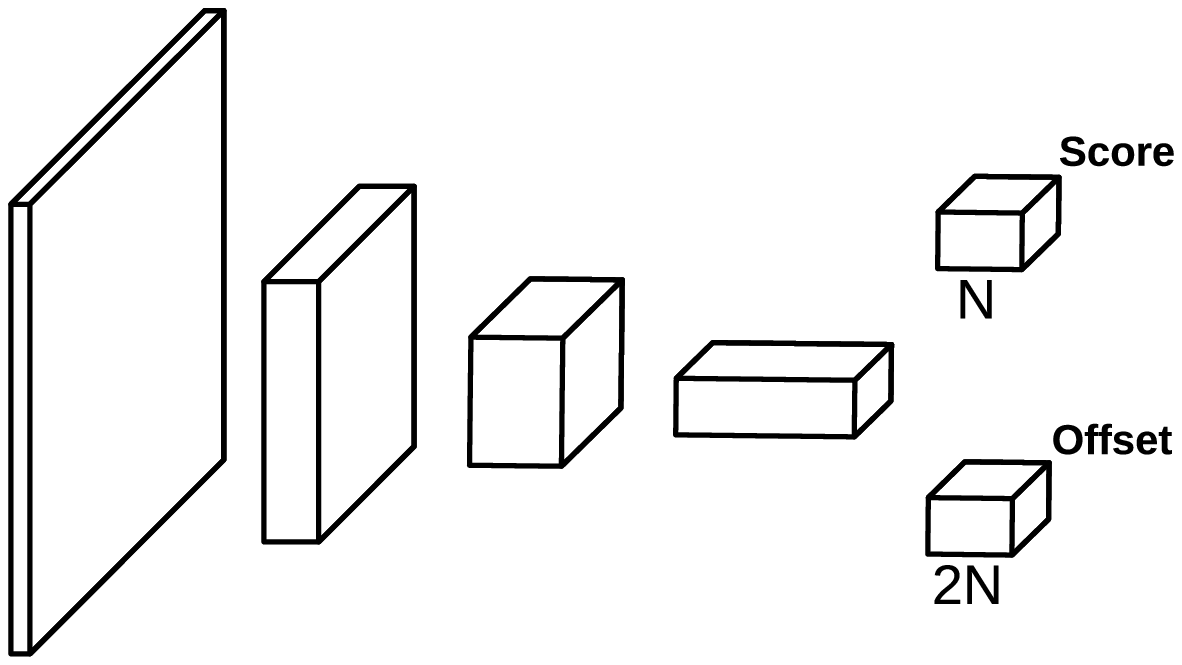}
    }
    \subfigure[PIP Regression + NRM\label{fig:heads_arc_4}]{
    \includegraphics[width=0.45\linewidth]{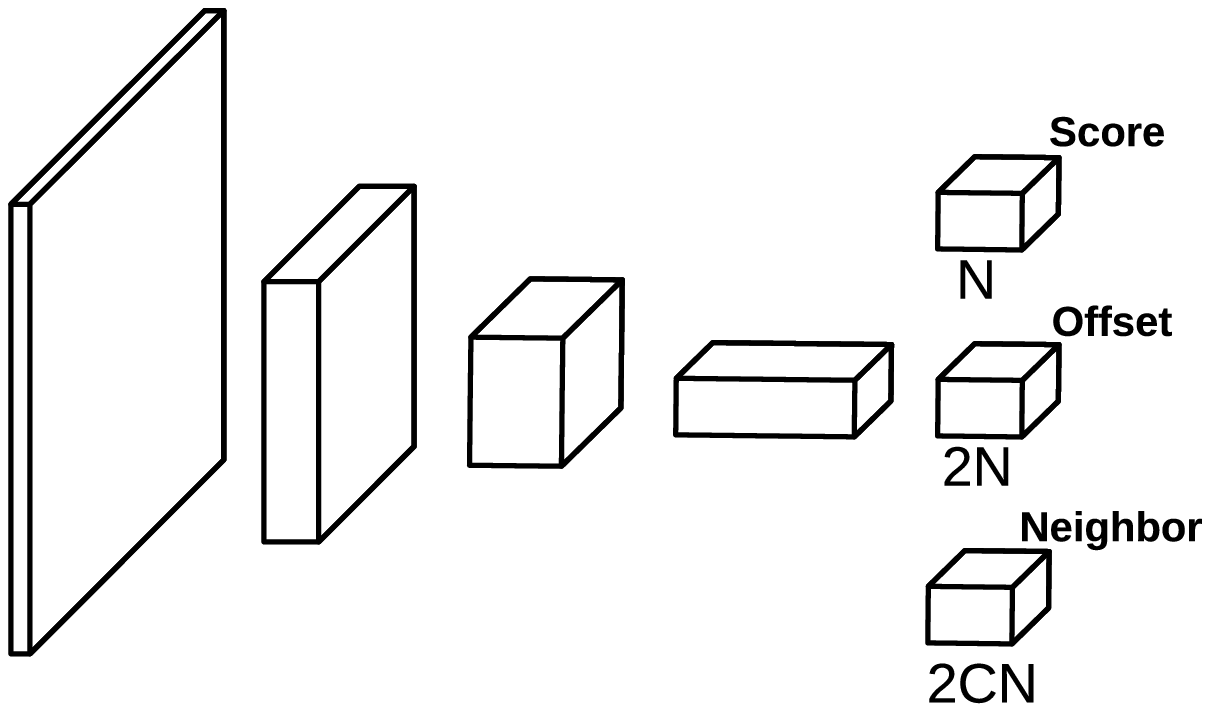}
    }
    \caption{Architectures of various detection heads. (a) Coordinate regression. (b) Heatmap regression. (c) PIP regression. (d) PIP regression + NRM.}
    \label{fig:heads_arc}
\end{figure}

\textbf{Heatmap Regression Models.} Heatmap regression maps an image to high-resolution heatmaps, each of which represents the probability of a landmark location. During inference, the location with the highest response on each heatmap is used. There are several ways to obtain high-resolution heatmaps. Stacked hourglass networks~\citep{NYD16,YLZ17,LZH19,CSJ19,DoY19,ZZY19,WBF19,CBG20} have been shown to perform well on landmark prediction through repeated downsampling and upsampling modules. U-Net~\citep{RFB15}, originally developed for biomedical image segmentation, has also been successfully applied to facial landmark detection~\citep{TPG18,ZZY19,DBC19,KMM20}. The convolutional pose machine (CPM)~\citep{WRK16,DYO18,DoY19} is a sequential architecture composed of CNNs, where the predictions are increasingly refined at each stage. \citet{RLZ19} use consecutive bilinear upsampling layers to recover high-resolution heatmaps. \citet{MRR18} maintain the input resolution through the whole network by not using any downsampling operations. \citet{XWW18} proposed a simple but effective architecture to obtain high-resolution heatmaps through several deconvolutional layers. High-Resolution Net (HRNet)~\citep{WSC19} maintains multi-resolution representations in parallel and exchanges information between these streams to obtain a final representation with great semantics and precise locations. However, existing works all require high-resolution heatmaps for heatmap regression, while PIP regression uses low-resolution heatmaps for reduced computational cost. 

The model most related to ours is from ~\citep{PZK17,PZC18}. This model predicts disk-shaped heatmaps as well as 2D offsets within the disk area, and the two predictions are then aggregated through Hough voting for person keypoint detection. Although this model and our proposed method can both be seen as hybrids of classifition and regression, there are considerable differences between the two. Firstly, our model is based on low-resolution feature maps, while the prior model uses high-resolution maps. Secondly, PIP regression (without NRM) itself is a hybrid of heatmap and coordinate regression, and also a general case of the two, while the prior model is essentially a heatmap regression model, whose offset prediction is an extra module for better accuracy. Finally, the neighbor regression module in this work aims to improve the consistency of predicted landmarks, while the short-range and mid-range offsets in ~\citep{PZC18} are mainly for accuracy improvement and keypoints grouping, respectively.

\textbf{Cross-Domain Generalization.} \citet{WQY18} introduced additional boundary information to help improve the robustness on unseen faces. \citet{ZSZ19} designed a geometry-aware module to address the occlusion problem. \citet{QSW19} proposed to augment the training data style with a conditional variational auto-encoder to enable models to generalize better on unseen domains. These methods all focus on improving cross-domain generalization through supervised learning, a paradigm which we call generalizable supervised learning. In contrast, we propose to address the cross-domain generalization issue for landmark detection through generalizable semi-supervised learning, enabling us to utilize massive amounts of unlabeled data.

\textbf{Semi-Supervised Facial Landmark Detection.} \citet{HMT18} proposed a module that can leverage unlabeled images by maintaining the consistency of predictions with respect to different image transformations. \citet{RLZ19} designed an adversarial training framework to leverage unlabeled data. \citet{DoY19} applied an interaction mechanism between a teacher and students in a self-training framework, where the teacher learns to estimate the quality of the pseudo-labels generated by the students. The key difference between the above methods and ours is that their labeled and test data are from the same domain, while we test on domains that do not contain any labeled (i.e., UDA) or even unlabeled data (i.e., GSSL). In other words, the focus of prior works is to improve the performance on the source domain with less labeled data, while we aim to obtain better generalization capability across domains, which is a more challenging task.

\begin{figure*}
\centering
    \subfigure[Input Image\label{fig:gt2label_1}]{
    \includegraphics[height=0.136\linewidth]{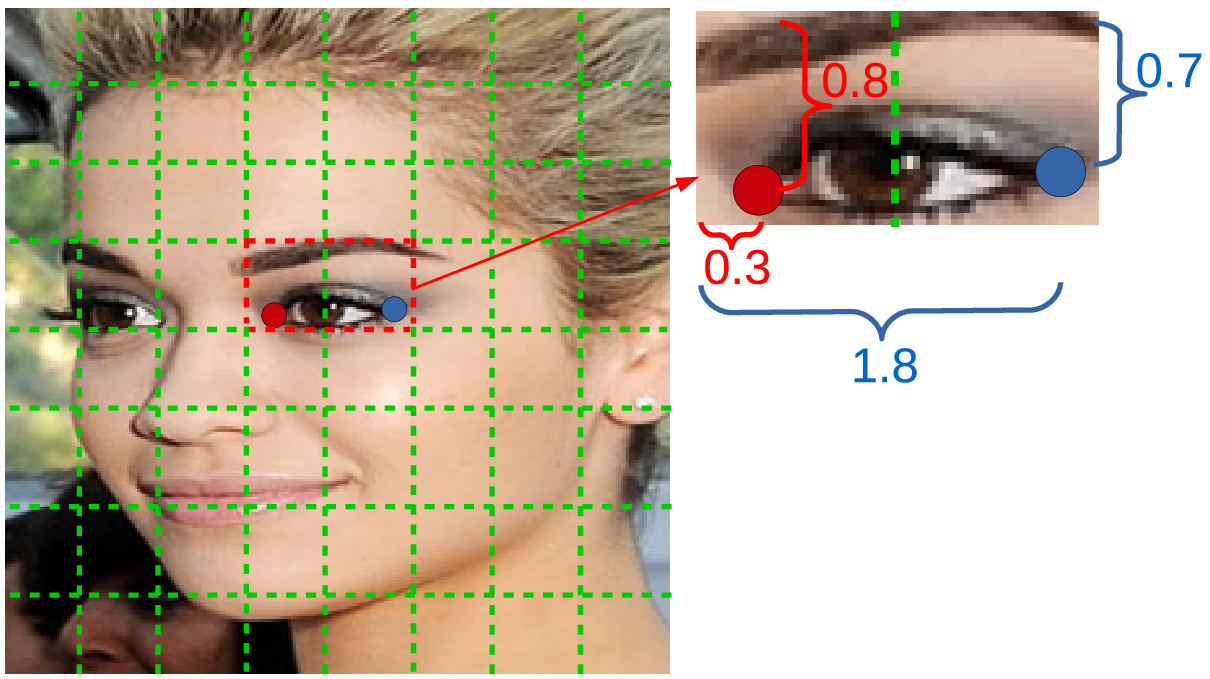}
    } 
    \subfigure[Score Map\label{fig:gt2label_2}]{
    \includegraphics[width=0.136\linewidth]{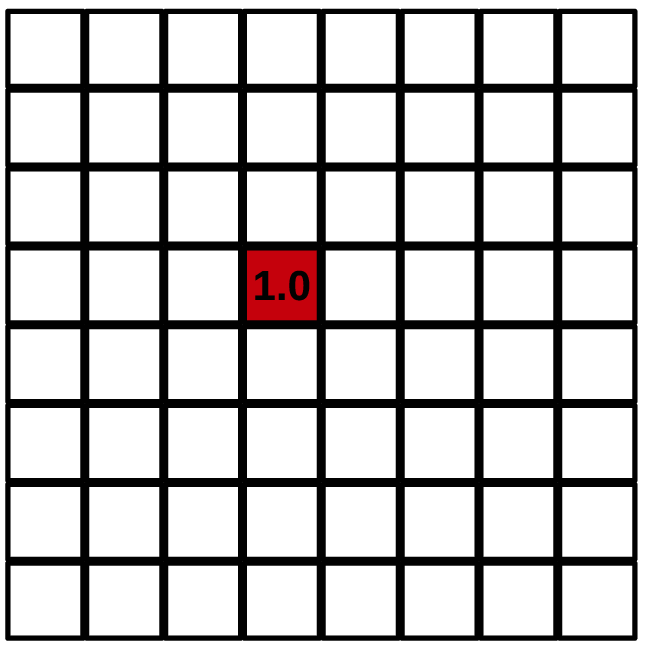}
    }     
    \subfigure[$x$-Offset Map\label{fig:gt2label_3}]{
    \includegraphics[width=0.136\linewidth]{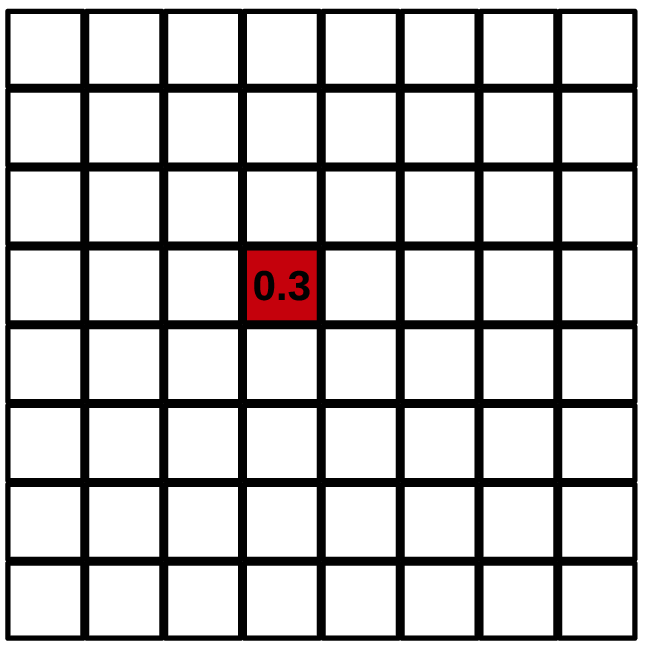}
    }
    \subfigure[$y$-Offset Map\label{fig:gt2label_4}]{
    \includegraphics[width=0.136\linewidth]{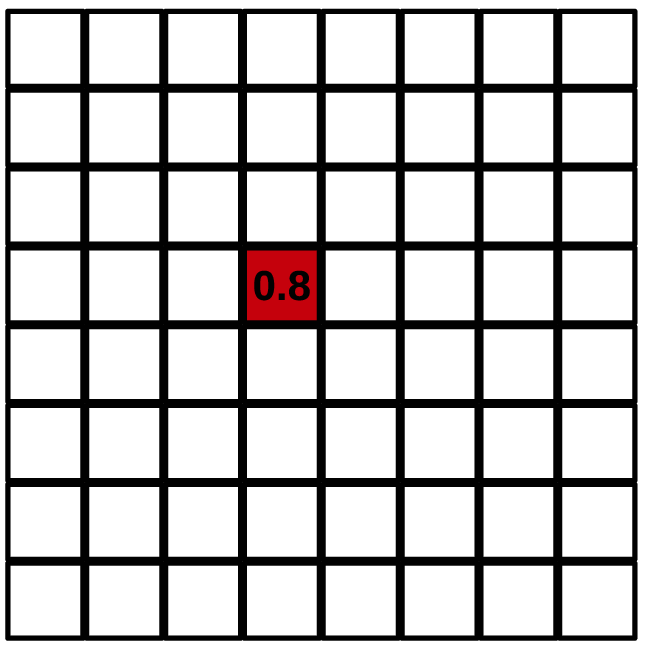}
    }        
    \subfigure[$x$-Neighbor Map\label{fig:gt2label_5}]{
    \includegraphics[width=0.136\linewidth]{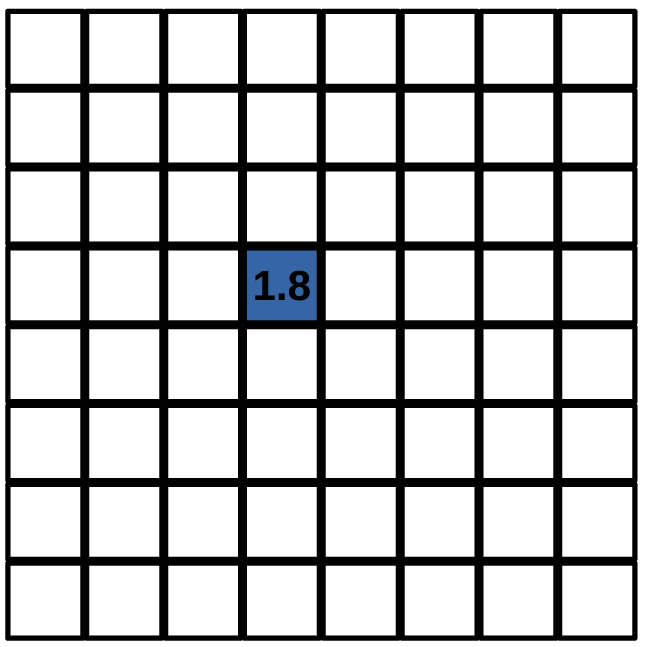}
    }
    \subfigure[$y$-Neighbor Map\label{fig:gt2label_6}]{
    \includegraphics[width=0.136\linewidth]{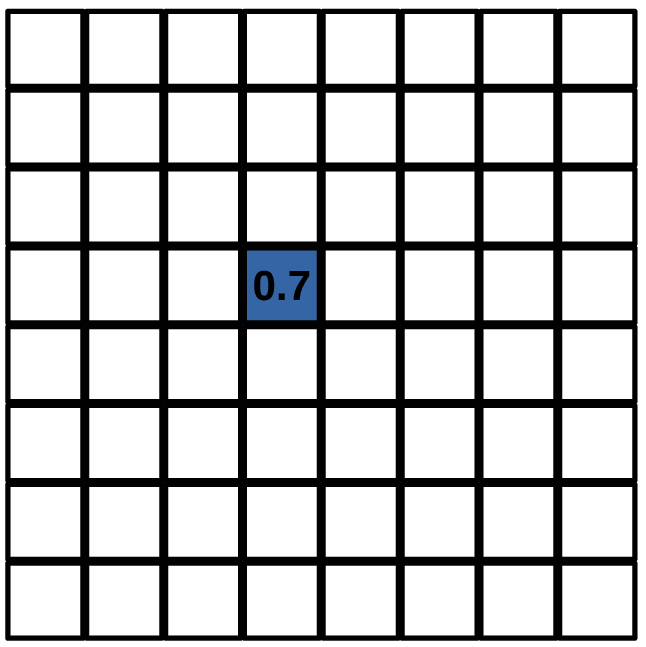}
    }
    \caption{Mapping from a ground-truth landmark to heatmap labels for PIPNet. (a) A sample image as input. The \textcolor{red}{red dot} denotes the target ground-truth landmark, and the \textcolor{blue}{blue one} is a neighboring landmark. (b) Label assignment for the score map. (c)-(d) Label assignment for the offset maps on $x$ and $y$ axes, respectively. (e)-(f) Label assignment for the neighbor maps on $x$ and $y$ axes, respectively.}
    \label{fig:gt2label}
\end{figure*}

\section{Our Method}
\label{sec:3}

In this section, we first introduce PIP regression (Section~\ref{sec:3.1}), and then present the proposed neighbor regression module (Section~\ref{sec:3.2}). We describe the self-training with curriculum framework  in Section~\ref{sec:3.3}. Finally, we present the implicit prior we observe from CNN-based facial landmark detectors in Section~\ref{sec:3.4}.   

\subsection{PIP Regression}
\label{sec:3.1}

Existing facial landmark detectors can be categorized into two classes, defined according to the type of detection head: coordinate regression and heatmap regression. As can be seen from Figure~\ref{fig:heads_arc_1}, coordinate regression outputs a vector with length $2N$ from fully connected layers, where $N$ represents the number of landmarks. Heatmap regression (see Figure~\ref{fig:heads_arc_2}), on the other hand, first gradually upsamples the extracted feature maps to the same (or similar) resolution as the input, and then outputs a heatmap with $N$ channels, each of which reflects the likelihood of the corresponding landmark location. When comparing the two detection heads, it is easy to see that coordinate regression is more computationally efficient at locating a point because heatmap regression needs to either upsample the feature maps repeatedly or maintain high-resolution feature maps throughout the network. However, heatmap regression has been shown to consistently outperform coordinate regression in terms of detection accuracy. Despite its inefficiency, heatmap regression is able to achieve state-of-the-art accuracy with a single-stage architecture, while coordinate regression usually needs two or more stages. As such, we pose the following question: Is it possible to obtain a detection head that is both efficient and accurate at the same time? 

We propose a novel detection head, termed PIP regression, which is built upon heatmap regression. We argue that upsampling layers are not necessary for locating points on feature maps. That is to say, low-resolution feature maps are sufficient for localization. By applying heatmap regression to low-resolution feature maps, we obtain the most likely grid on the heatmap for each landmark. To get more precise predictions, we also apply offset prediction within each heatmap grid on the $x$-axis and $y$-axis, relative to the top-left corner of the grid. It is worth noting that PIP regression is a single-stage method because the score and offset predictions are independent to each other, and can thus be computed in parallel. Figure~\ref{fig:heads_arc_3} gives the architecture of PIP regression, where the outputs are a score map ($N \times H_{M} \times W_{M}$) and an offset map ($2N \times H_{M} \times W_{M}$). The proposed detection head can be simply implemented by a $1\times1$ convolutional layer.

Figure~\ref{fig:gt2label} demonstrates how to convert a ground-truth landmark to heatmap labels for PIPNet. Suppose Figure~\ref{fig:gt2label_1} is an input image of size $256 \times 256$, the red dot on the right inner-eye-corner is the ground-truth landmark, and the network stride is 32. Then, the last feature map is of size $8 \times 8$. As can be seen from the figure, there are 64 grids on the last feature map for each channel, and we denote the grid that the ground-truth falls into as the positive grid. For the score map (see Figure~\ref{fig:gt2label_2}), the positive grid is assigned 1, and the rest are 0. Because the ground-truth landmark has a 30\% offset on the $x$-axis relative to the top-left corner of the positive grid, the positive grid on the $x$-offset map is assigned 0.3 (see Figure~\ref{fig:gt2label_3}). Similarly, the same grid on the $y$-offset map is assigned 0.8 (see Figure~\ref{fig:gt2label_4}), and the rest are 0. The training loss for PIP regression can be formulated as follows:

\begin{figure*}
\centering
    \subfigure[Coordinate Regression (Avg. NME=\textbf{6.61})\label{fig:WFLW_vis_coord}]{
    \includegraphics[width=1\linewidth]{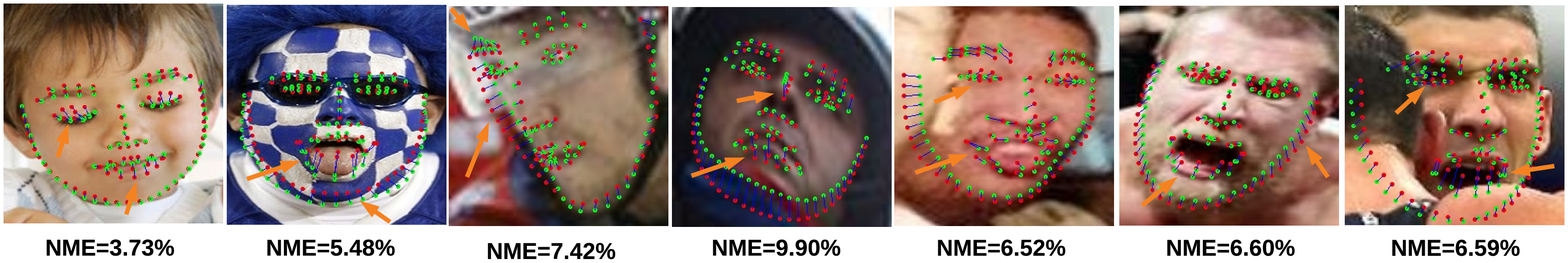}
    } 
    
    \subfigure[Heatmap Regression, Stride=1 (Avg. NME=\textbf{5.90})\label{fig:WFLW_vis_map}]{
    \includegraphics[width=1\linewidth]{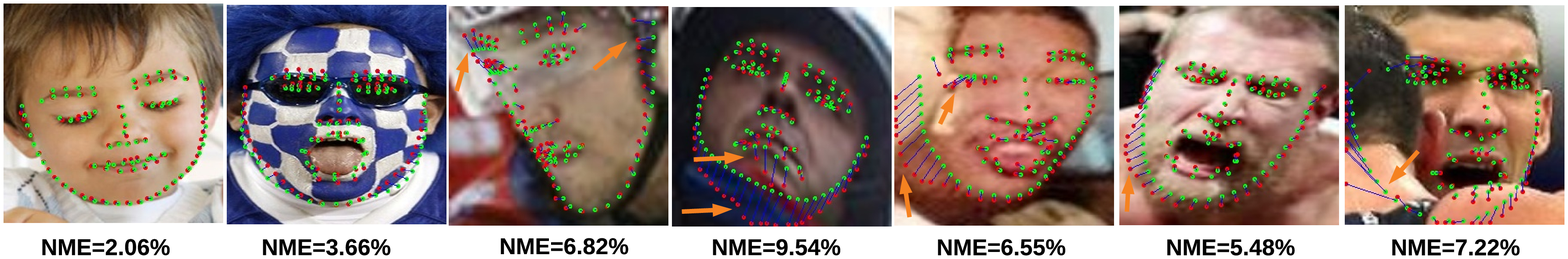}
    }     
    
    \subfigure[PIP Regression, Stride=32 (Avg. NME=\textbf{4.84})\label{fig:WFLW_vis_pip}]{
    \includegraphics[width=1\linewidth]{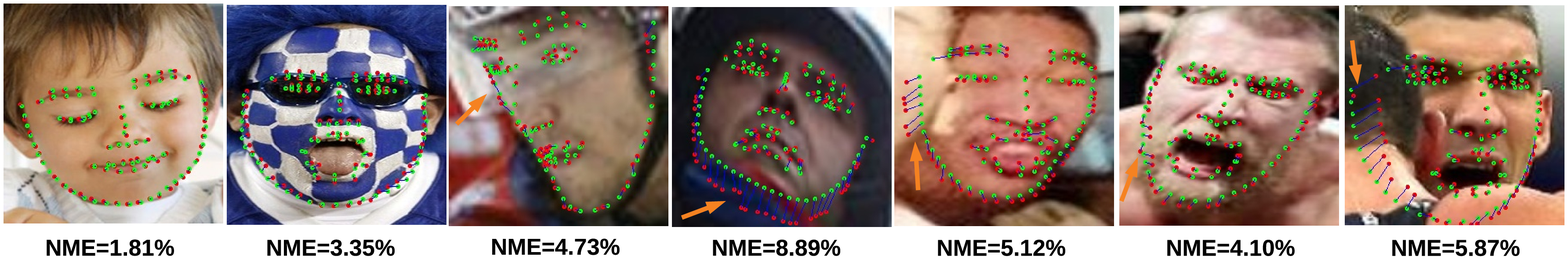}
    }
    
    \subfigure[PIP Regression + NRM, Stride=32 (Avg. NME=\textbf{4.67})\label{fig:WFLW_vis_pip_nrm}]{
    \includegraphics[width=1\linewidth]{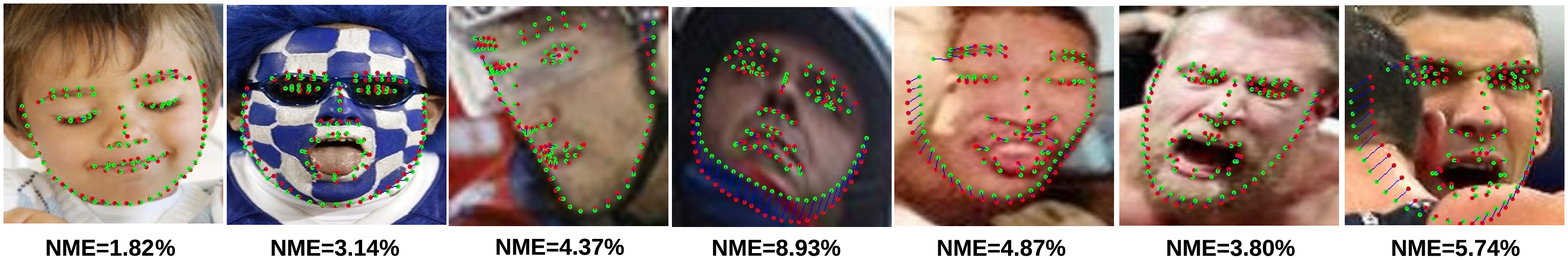}
    }            
    \caption{Visualization of predicted results on sample images from WFLW test set, with different detection heads. \textcolor{green}{Green dots} are ground-truths, \textcolor{red}{red ones} are predictions, and \textcolor{orange}{orange arrows} show the areas with bad predictions. (a) Predictions of coordinate regression. (b) Predictions of heatmap regression, with stride 1. (c) Predictions of PIP regression, with stride 32. (d) Predictions of PIP regression + NRM, with stride 32. \label{fig:WFLW_vis}}
\end{figure*}

\begin{equation} 
\label{eq:1}
L =  L_{S} + \alpha L_{O},
\end{equation}
where $L_{S}$ is the loss for score prediction, $L_{O}$ is for offset prediction, and $\alpha$ is a balancing coefficient. Concretely, $L_{S}$ for a score map ($N \times H_{M} \times W_{M}$) is formulated as 

\begin{equation}
\label{eq:2}
L_{S} = \frac{1}{NH_{M}W_{M}} \sum_{i=1}^{N} \sum_{j=1}^{H_{M}} \sum_{k=1}^{W_{M}} (s_{ijk}^\ast - s_{ijk}^\prime)^2 , \quad s_{ijk}^\ast \in \lbrace 0,1 \rbrace,
\end{equation}
where $s_{ijk}^\ast$ and $s_{ijk}^\prime$ denote the ground-truth and predicted score values, respectively, and $NH_{M}W_{M}$ is the normalization term. $L_{O}$ for an offset map ($2N \times H_{M} \times W_{M}$) is formulated as

\begin{equation}
\label{eq:3}
L_{O} = \frac{1}{2N} \sum_{s_{ijk}^\ast=1} \sum_{l=1}^2 |o_{ijkl}^\ast - o_{ijkl}^\prime|,  \quad o_{ijkl}^\ast \in [0,1],
\end{equation}
where $o_{ijkl}^\ast$ and $o_{ijkl}^\prime$ denote the ground-truth and predicted offset values, respectively, and $2N$ is the normalization term. As can be seen from Equation~\ref{eq:2} and \ref{eq:3}, $L_{S}$ is applied to all the samples of a score map, while $L_{O}$ is applied to only positive samples of an offset map. Moreover, we use different loss fuctions for $L_{S}$ and $L_{O}$ because the former is actually a classification problem, while the latter is a regression problem. According to~\citep{FKA18}, the L1 loss yields better results for regression, which is consistent with our experimental results. On the other hand, our experiments indicate that the L2 loss is better for classification. Therefore, we use the L2 loss for $L_{S}$ and the L1 loss for $L_{O}$. During inference, the final prediction of a landmark is computed as the grid location with the highest response refined by its corresponding offsets.  

\begin{figure*}
\centering
    \subfigure[\label{fig:WFLW_vis_detail_c1}]{
    \includegraphics[width=0.485\linewidth]{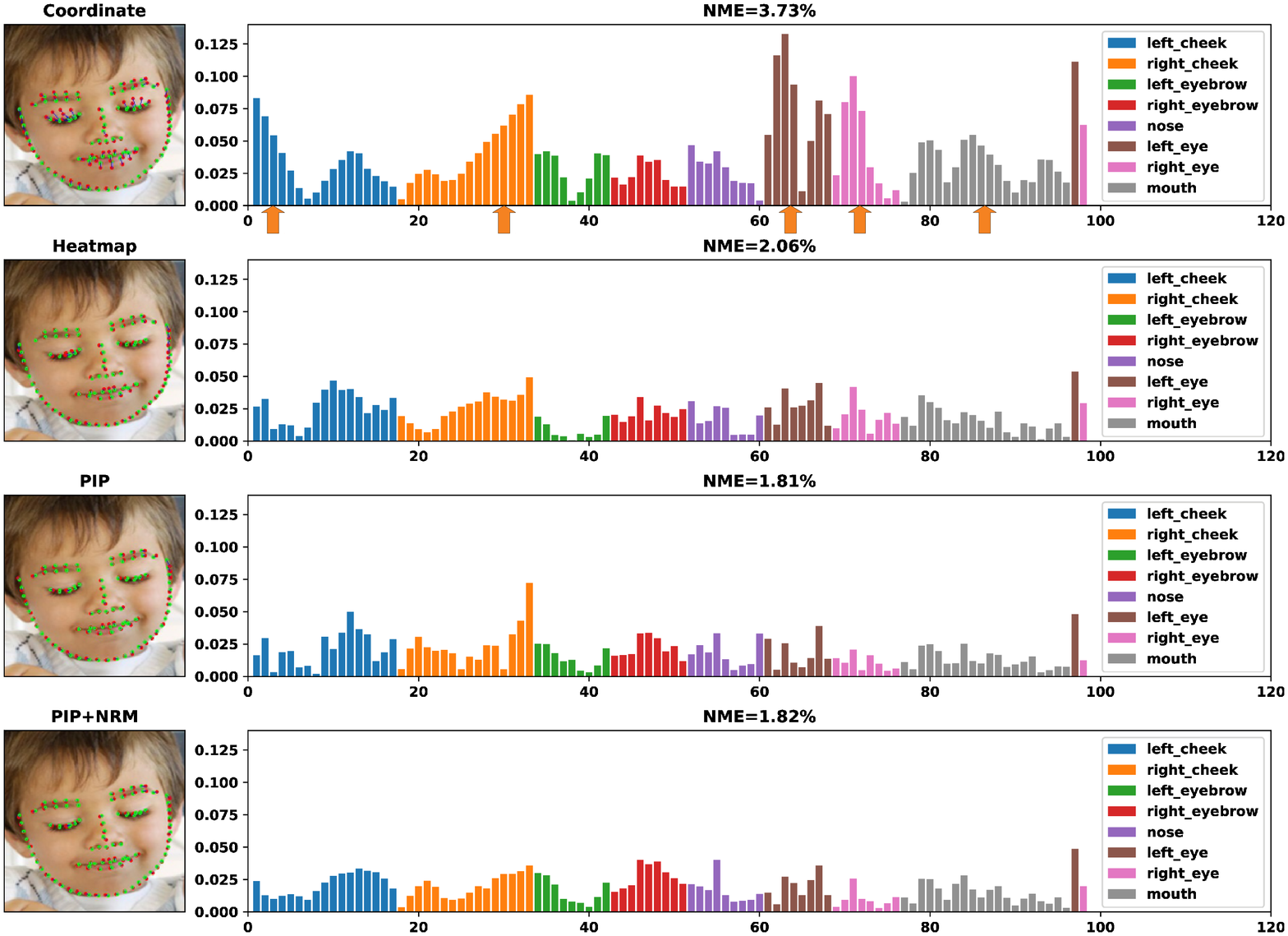}
    }    
    \subfigure[\label{fig:WFLW_vis_detail_c6}]{
    \includegraphics[width=0.485\linewidth]{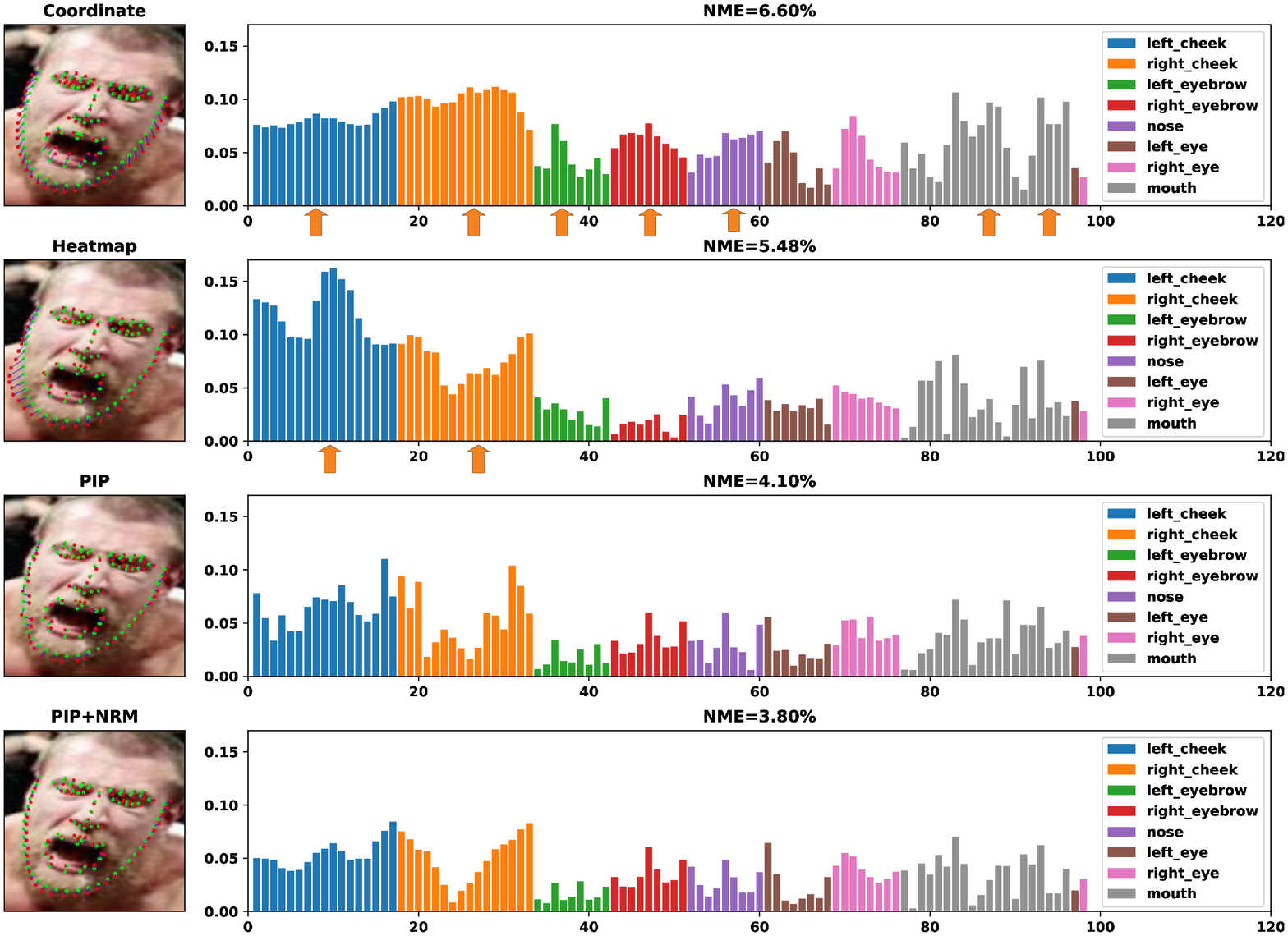}
    }                
    \caption{Bar charts of normalized error on each landmark. The landmark IDs follow the original annotation, and are grouped into eight categories according to their regions for easy observation. \textcolor{orange}{Orange arrows} show the areas with relatively large errors. (a) Images from 1st column of Figure~\ref{fig:WFLW_vis}. (b) Images from 6th column of Figure~\ref{fig:WFLW_vis}. \label{fig:WFLW_vis_detail}}
\end{figure*}

One hyperparameter of PIP regression is the stride of the network. Given the image size and network stride, the size of the heatmap can be determined as follows.

\begin{equation}
\label{eq:4}
H_{M} = \frac{H_{I}}{S}, \quad W_{M} = \frac{W_{I}}{S},
\end{equation}
where $H_{I}$ and $W_{I}$ are the height and width of the input image, and $S$ denotes the network stride. Intuitively, PIP regression can be seen as a generalization of the two existing detection heads. When the network stride is equal to the image size (i.e., $H_{M}=W_{M}=1$), and the score prediction module is removed, PIP regression can be seen as coordinate regression, where the conventional fully connected layers are replaced by convolutional layers. When the network stride is equal or close to $1$, and the offset prediction is removed, then PIP regression is equivalent to heatmap regression (though there are still differences in implementation details, such as label smoothing and landmark inference). Consequently, PIP regression can be seen as conducting heatmap regression globally and coordinate regression locally at the same time, which is the reason it is called 'pixel-in-pixel'. Such a property endows PIP regression with better flexibility and greater potential than the two alternatives.

\subsection{Neighbor Regression Module}
\label{sec:3.2}

Although the proposed PIP regression addresses the computational efficiency issue of heatmap regression, it still suffers from poor robustness. Figure~\ref{fig:WFLW_vis_coord}-\ref{fig:WFLW_vis_pip} show some sample images with predictions from coordinate regression, heatmap regression, and PIP regression, respectively. As can be seen from Figure~\ref{fig:WFLW_vis_coord}, coordinate regression outputs predictions with reasonable global shapes, even on large poses (e.g., 4th, 5th, and 7th images). However, it is not accurate in details, as we can observe obvious shifts between predictions and ground-truths in some areas (e.g., \textit{eye} and \textit{mouth} areas of the 1st image, \textit{mouth} of the 2nd image, \textit{left cheek} of the 3rd image, etc.). Consequently, coordinate regression may not be able to detect subtle changes such as eye blinking and mouth opening, which are essential functions of anti-spoofing. In contrast, as shown in Figure~\ref{fig:WFLW_vis_map}, heatmap regression is precise in details in general, but can give inconsistent shapes for images with extreme poses (e.g., 4th to 7th images). Similarly, PIP regression can also lack robustness under extreme poses (see 4th to 7th images of Figure~\ref{fig:WFLW_vis_pip}), despite obtaining better normalized mean error (NME) than heatmap regression. It is not difficult to understand the lack of robustness in heatmap-based models, because their predictions are based on different features (i.e., different locations on the feature map) and are thus independent to each other. In contrast, all the landmarks predicted by coordinate regression share the same feature, which we believe is the key to robustness. 

Inspired by the above findings, we further propose a neighbor regression module (NRM) to help PIP regression predict more consistent landmarks. Specifically, in addition to the offsets of the landmark itself, each landmark also predicts the offsets of its $C$ neighbors. As shown in Figure~\ref{fig:heads_arc_4}, NRM further outputs a neighbor map of size $2CN \times H_{M} \times W_{M}$, where $C$ is the number of neighbors to predict. Concretely, mean shapes of the face landmarks are computed using the ground-truths in the training data, and the $C$ closest landmarks of the target landmark (excluding itself) are defined as its neighbors. In this work, we simply use Euclidean distance as the distance metric. We also explored correlation for defining neighbors, which gives a similar performance to Euclidean distance. Figure~\ref{fig:gt2label_5}-\ref{fig:gt2label_6} describe the label assignment for the neighbor maps of the ground-truth landmark (i.e., the red dot). Here, we use only one neighbor for illustration, but there can be more than one in practice. Assume that the blue dot on the right outer-eye-corner is a neighbor of the red dot in Figure~\ref{fig:gt2label_1}. As can be seen from the figure, the blue dot has 180\% and 70\% offset on $x$ and $y$ axes respectively, relative to the top-left corner of the positive grid of the red dot. Thus, we assign 1.8 and 0.7 to the positive grids on $x$- and $y$-neighbor maps, respectively, and the remaining grids are 0. Note that the score, offset, and neighbor maps all belong to the red dot, while the blue dot has its own maps, which are not shown here. 

After adding NRM, the training loss of PIPNet becomes 

\begin{equation}
\label{eq:5}
L = L_{S} + \alpha L_{O} + \beta L_{N},
\end{equation}
where $L_{N}$ is the loss for NRM, and $\beta$ is another balancing coefficient. We define $L_{N}$ as follows.

\begin{equation}
\label{eq:6}
L_{N} = \frac{1}{2CN} \sum_{s_{ijk}^\ast=1} \sum_{l=1}^{2} \sum_{m=1}^{C} |n_{ijklm}^\ast - n_{ijklm}^\prime|,  \quad n_{ijklm}^\ast \in [0,1],
\end{equation}
where $n_{ijklm}^\ast$ and $n_{ijklm}^\prime$ denote the ground-truth and predicted neighbor offset values, respectively, and $2CN$ is the normalization term. Like $L_{O}$, $L_{N}$ also uses the L1 loss because it is a regression problem. During inference, each landmark collects its locations predicted by other landmarks as well as its own prediction, and then calculates the average of these as its final prediction. 

\begin{figure}
\centering
  \includegraphics[width=0.87\linewidth]{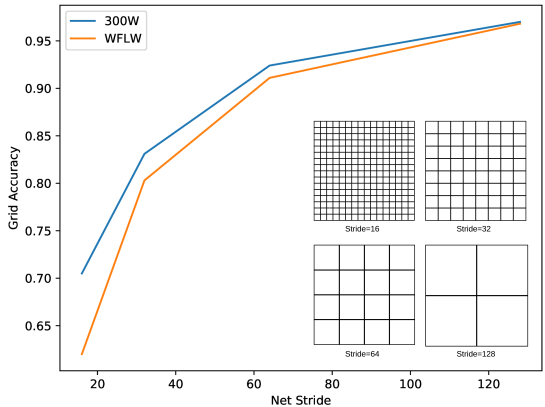}
\caption{Classification accuracy on grids vs. network stride, tested on 300W and WFLW. The bottom-right of the figure gives the visualized heatmaps with different strides to better understand the difference in classification difficulty.}
\label{fig:map_acc}       
\end{figure}

With the help of NRM, PIP regression becomes more robust in addition to being accurate, as evidenced by Figure~\ref{fig:WFLW_vis_pip_nrm}. Since the small errors are not easy to perceive through images, we further transform the errors to bar charts for a clearer illustration. The bar charts of two clolumns (i.e., 1st and 6th) from Figure~\ref{fig:WFLW_vis} are presented in Figure~\ref{fig:WFLW_vis_detail_c1} and \ref{fig:WFLW_vis_detail_c6} respectively, where each bar represents the normalized error of a landmark and the landmark IDs follow the original annotation (see supplementary material for landmark IDs and the other columns). To make it easy for observation, the bars are grouped into eight categories according to their regions. Although the image in Figure~\ref{fig:WFLW_vis_detail_c1} is relatively simple, the predictions of coordinate regression have obvious shifts at \textit{left and right cheek}, \textit{left and right eye}, and \textit{mouth} areas, which is the reason it obtains the worst NME=3.73\%. In contrast, heatmap-based methods all achieve small NMEs thanks to their advantage on local accuracy. For the image in Figure~\ref{fig:WFLW_vis_detail_c6}, it is more difficult due to the expression and blurring issue. From the bar charts in Figure~\ref{fig:WFLW_vis_detail_c6}, we see that coordinate regression has large NME due to its accumulated local errors at multiple regions such as \textit{left and right eyebrow}, \textit{nose}, and \textit{mouth}, despite its satisfactory global shape. Heatmap regression, on the contrary, obtains a large NME because of the inconsistent predictions at \textit{left cheek} area. Being a heatmap-based method, PIP regression also predicts inconsistent landmarks on \textit{left cheek} due to blurring and unclear boundaries. By adding NRM, PIP regression improves on both qualitative (consistency of predictions) and quantitative (NME value) results. Please refer to Section~\ref{sec:4.2.2} and \ref{sec:4.3.2} for more quantitative results on NRM.

\begin{figure}
\centering
  \includegraphics[width=0.9\linewidth]{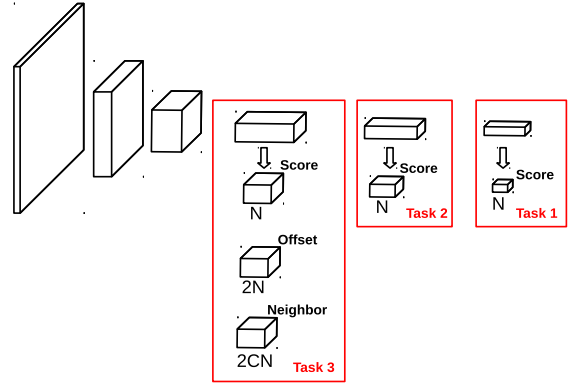}
\caption{Architecture of PIPNet with the STC strategy. Two more higher-stride heatmap regression layers are added after PIP regression layer.}
\label{fig:stc_arc}     
\end{figure}

\subsection{Self-Training with Curriculum}
\label{sec:3.3}

Although the neighbor regression module alleviates the unstableness of PIP regression, it is still not adequate for cross-domain datasets. \citet{VBV19} pointed out the existence of domain gaps in facial landmark detection, and our experiments in Section~\ref{sec:4.3.2} also confirm this problem. Therefore, we would like to further utilize unlabeled data across domains to improve the cross-domain generalization capability of our model. To this end, we propose self-training with curriculum (STC), which is built upon self-training. The key difference between traditional self-training and our method is that in the former the task is fixed, while in our method the difficulty of the task gradually increases, mimicking how humans learn. Different from original curriculum learning~\citep{BLC09}, which presents training examples progressively for a specific task, our strategy applies curriculum learning at a task level. Such a design is based on the observation that grid classification on heatmaps becomes easier when the stride of the network becomes larger. This is easy to understand because a lower-resolution heatmap has less negative grids. Figure~\ref{fig:map_acc} shows the change in classification accuracy on grids when the network stride varies, tested on two datasets. It is clear that the classification accuracy increases consistently as the network stride becomes larger. We also observe that the advantage is more obvious for large network strides on harder datasets (WFLW contains more in-the-wild images than 300W), which indicates that the strategy will be more effective on harder unlabeled datasets. In the bottom-right corner of the figure, we give the heatmaps under different strides to better understand the differences in difficulty of classification.

Thanks to the flexibility of PIP regression, a PIPNet for supervised learning can easily be converted to a model with STC by simply adding higher-stride heatmap regression layers on top of the PIP regression. Figure~\ref{fig:stc_arc} gives the architecture of PIPNet with the STC strategy. As can be seen, two more heatmap regression layers are added to a standard PIPNet. Assume that the input image is of size $256\times256$, and the standard PIP regression is of stride 32. Then, the heatmap size of the PIP regression layer is $8\times8$, and the sizes of the added ones are $4\times4$ and $2\times2$. In conventional self-training, the model iteratively learns from pseudo-labeled images on a fixed task (in our case, \textit{Task 3} in the figure) until it converges. In contrast, in the proposed STC, the sequence of the tasks is arranged as \textit{Task 1} $\rightarrow$ \textit{Task 2} $\rightarrow$ \textit{Task 3} $\circlearrowleft$, where the difficulty gradually increases until \textit{Task 3}. By doing so, less errors from pseudo-labels are introduced and learned by the model so that the mistake reinforcement problem of self-training can be eased.

\begin{figure}
\centering
    \subfigure[Plain black image\label{fig:prior_1}]{
    \includegraphics[width=0.47\linewidth]{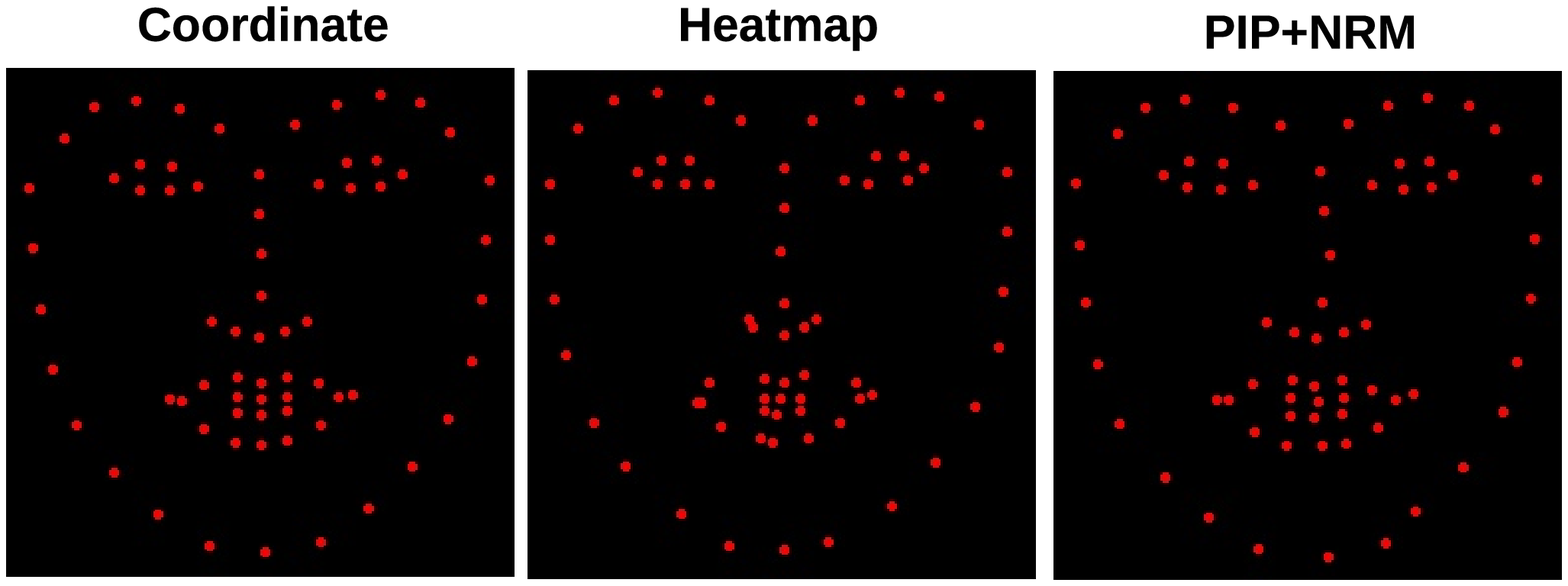}
    }     
    \subfigure[CIFAR-10: cat\label{fig:prior_2}]{
    \includegraphics[width=0.47\linewidth]{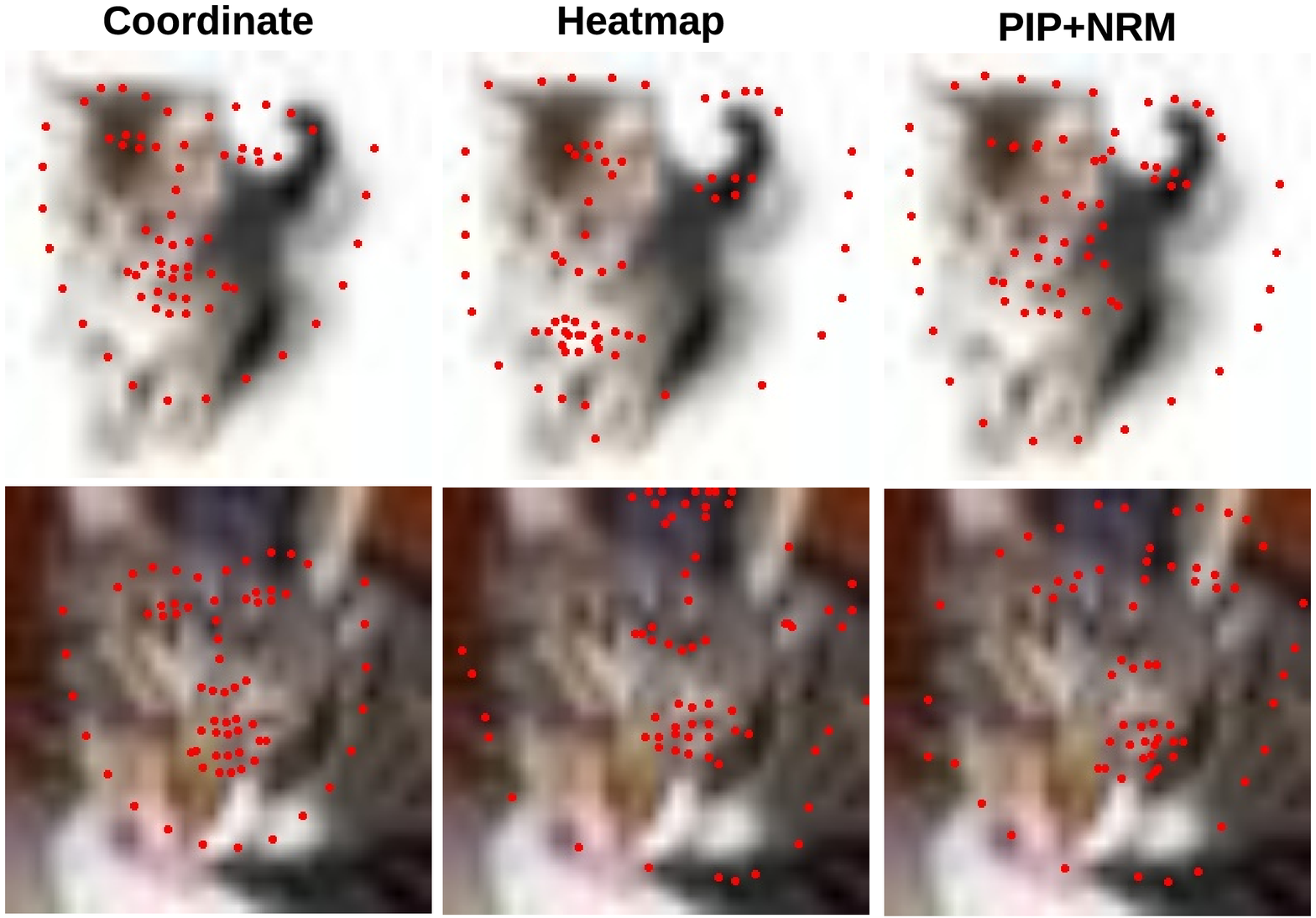}
    }       
      
    \subfigure[CIFAR-10: deer\label{fig:prior_3}]{
    \includegraphics[width=0.47\linewidth]{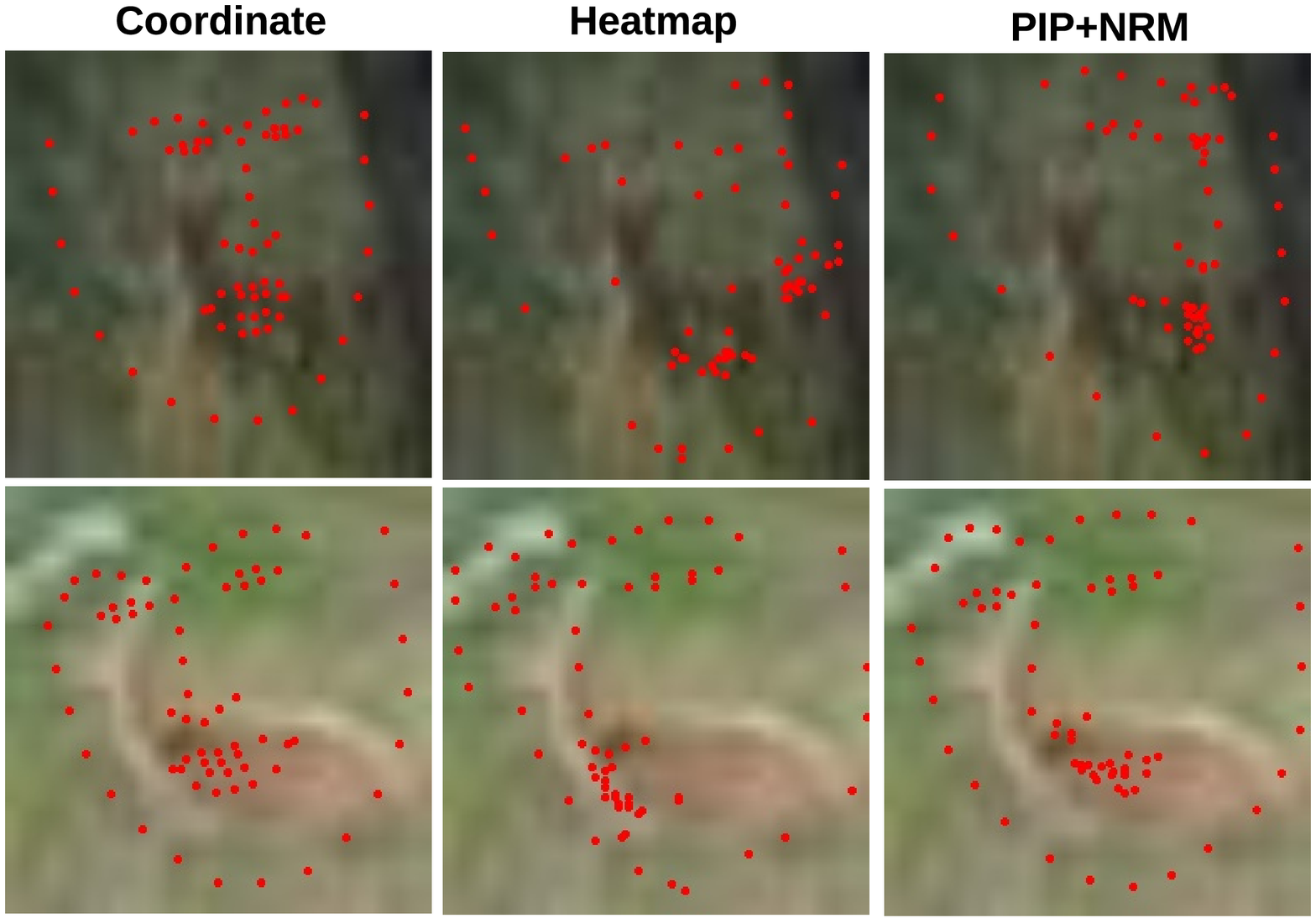}
    }   
    \subfigure[CIFAR-10: truck\label{fig:prior_4}]{
    \includegraphics[width=0.47\linewidth]{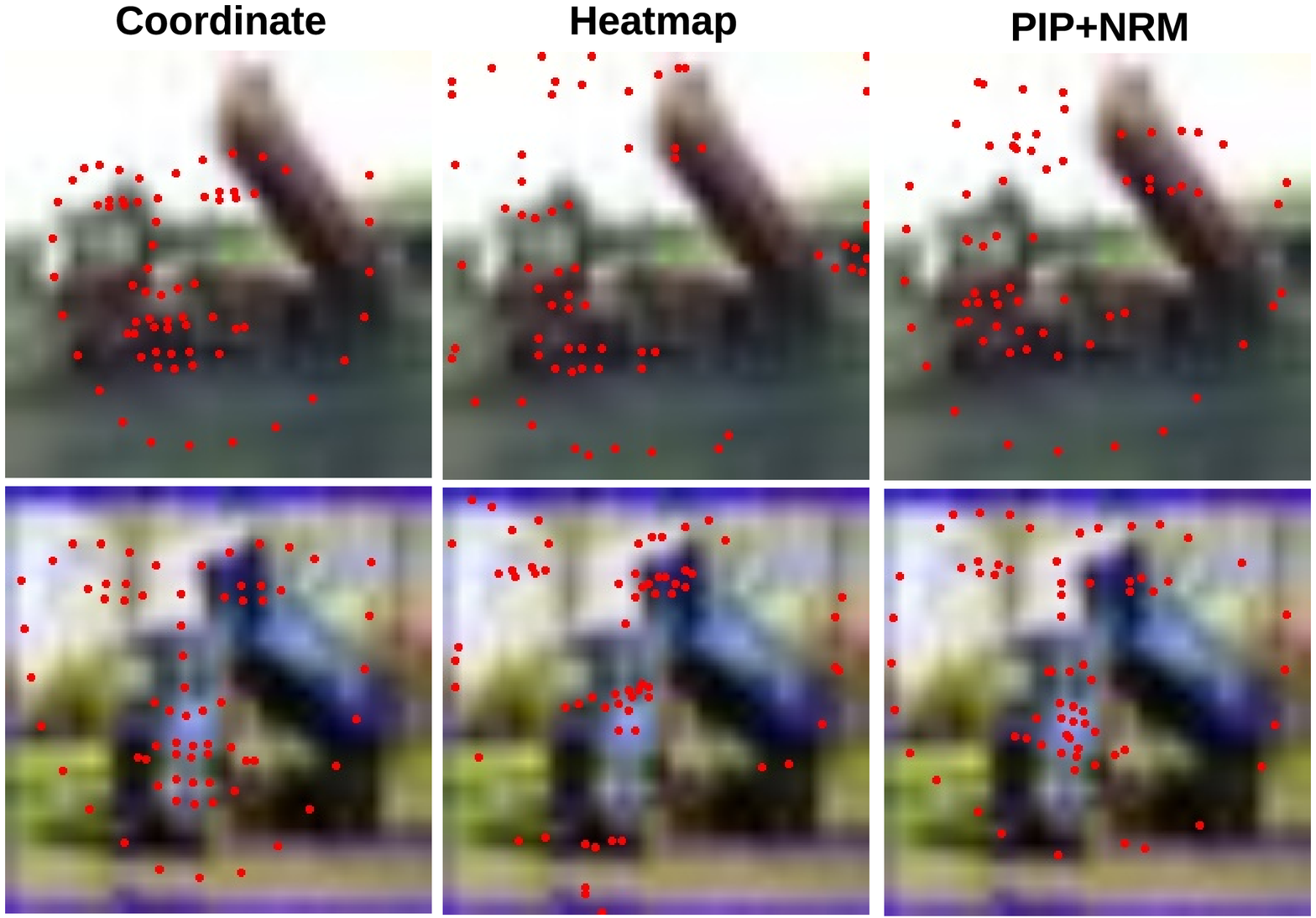}
    }            
    \caption{Predictions from CNN-based landmark detectors, tested with three detection heads: coordinate regression, heatmap regression, and PIP regression + NRM. (a) Trained on plain black images but normal ground-truths from 300W, tested on plain black images. (b)-(d) Trained on 300W normally, tested on images from CIFAR-10. \label{fig:prior}}
\end{figure}

The pipeline of self-training with curriculum can be simply described as follows: (1) The modified PIPNet is trained with manually labeled data in a standard way (i.e., \textit{Task 3}); (2) Pseudo-labels of the unlabeled data are estimated using the trained detector; (3) A new training set is formed with the manually labeled and pseudo-labeled data. (4) The modified PIPNet is trained on the new training set, using the manually labeled data to train the model through \textit{Task 3}, and the pseudo-labeled data through \textit{Task X}. Steps (2) to (4) are repeated until the model converges, and \textit{Task X} is selected sequentially from the sequence \textit{Task 1} $\rightarrow$ \textit{Task 2} $\rightarrow$ \textit{Task 3} $\circlearrowleft$. Empirically, we find that the model converges after three iterations of \textit{Task 3}, which is used in all the relevant experiments. During inference, the model is used in the same way as the standard PIPNet, and the added heatmap regression layers are simply ignored or discarded.

\subsection{Implicit Prior}
\label{sec:3.4}

\begin{figure}
  \includegraphics[width=1\linewidth]{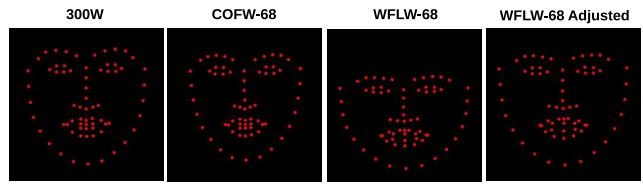}
\caption{Mean faces of 300W, COFW-68, WFLW-68, and WFLW-68 after bounding box adjustment.}
\label{fig:wflw_adjust}       
\end{figure}

As pointed out by \citet{IJB20}, a CNN is able to encode position information through zero paddings. In other words, the neurons of a CNN know which part of an image they are looking at. To verify this, we train facial landmark detectors using all plain black images, but the ground-truth landmarks remain unchanged. Then, we input a plain black image for testing. The predictions of three different detection heads are shown in Figure~\ref{fig:prior_1}. As can be seen from the figure, the models memorize the most likely positions of the landmarks regradless of which detection head they use, which proves their ability to perceive absolute positions. Therefore, CNNs do learn what (semantic features) and where (absolute positions) jointly~\citep{IJB20}. Different from multi-person keypoint detection and general object detection, facial landmark detectors locate landmarks through a cropped face image, where the facial features are correlated to certain positions (despite the use of augmentation techniques, such as translation and rotation, during training). To validate this, we train models with normal face images but test on images without human faces. As shown in Figure~\ref{fig:prior_2}-~\ref{fig:prior_4}, when the input images come from CIFAR-10, the models still give landmark predictions close to a human face even if there is no facial feature information. That is to say, position information also contributes to the response of heatmaps. Intuitively, this can be seen as a prior implicitly learned by CNNs from training data. We also observe that coordinate regression gives more consistent predictions than the other two detection heads when no face is shown, which indicates that coordinate regression has a stronger prior of landmark positions. This may explain why coordinate regression is more robust but biased.

Another thing the implicit prior tells us is that it is important to have consistent cropped face images. While this may not be a problem in practice because the faces are usually detected by the same face detector, the benchmark datasets for facial landmark detection provide bounding boxes in different styles (see Figure~\ref{fig:wflw_adjust}). When conducting cross-domain evaluation or domain adaptation, bounding box styles should be consistent to avoid causing performance degradation. In our case, we reduce the top area of the bounding boxes provided in WFLW-68 by 20\% for the cross-domain setting. Figure~\ref{fig:wflw_adjust} presents the mean faces of 300W, COFW-68, and WFLW-68 (before and after adjustment). The 300W and COFW-68 datasets have similar cropping styles, but WFLW-68 is shifted significantly downward. After adjustment, the mean face of WFLW-68 is roughly aligned with those of 300W and COFW-68.

\section{Experiments}
\label{sec:4}

We first introduce the experimental settings in Section~\ref{sec:4.1}, and then discuss the hyperparameters of PIPNet in Section~\ref{sec:4.2}. In Section~\ref{sec:4.3}, we analyze the characteristics of PIPNet by comparing it to the baselines. We present the performance of PIPNet under the supervised and cross-domain setting in Sections~\ref{sec:4.4} and \ref{sec:4.5}, respectively. Finally, we demonstrate the advantage of PIPNet in terms of inference speed (Section~\ref{sec:4.6}).

\subsection{Experimental Settings}
\label{sec:4.1}

\subsubsection{Datasets}
\label{sec:4.1.1}

\textbf{300W}~\citep{STZ13} provides 68 landmarks for each face in images collected from LFPW, AFW, HELEN, XM2VTS, and IBUG. Following~\citep{RCW16}, the 3,148 training images come from the training sets of LFPW and HELEN, and the full set of AFW. The 689 test images are from the test set of LFPW and HELEN, and the full set of IBUG.  The test images are further divided into two sets: the common set (554 images) and the challenging set (135 images). Note that the common set is from LFPW and HELEN, and the challenging set is from IBUG.

\textbf{COFW}~\citep{BPD13} contains 1,345 training images and 507 test images, with the face images having large variations and occlusions. Originally, 29 landmarks were provided for each face. \citet{GhF14} reannotated the test set with 68 landmarks, which we denote as COFW-68. We use the original annotations for the supervised setting and the 68-landmark version for the cross-domain setting.

\textbf{WFLW}~\citep{WQY18} consists of 7,500 training images and 2,500 test images from WIDER Face~\citep{YLL16}, with each face having 98 annotated landmarks. The faces in WFLW introduce large variations in pose, expression, and occlusion. The test set is further divided into six subsets for a detailed evaluation. These include pose (326 images), expression (314 images), illumination (698 images), make-up (206 images), occlusion (736 images), and blur (773 images). The original annotations are used under the supervised setting. To make WFLW applicable for the cross-domain setting, we generate 68-landmark annotations for the test set by converting the original 98 landmarks, and we name the new annotated dataset WFLW-68. Please refer to the supplementary materials for more details on the conversion.   

\textbf{AFLW}~\citep{KWR11} contains 24,386 face images in total, 20,000 of which are training images, with the remaining 4,386 used for testing. Following~\citep{ZLL16a}, we use 19 landmarks of AFLW for training and testing.

\textbf{Menpo 2D}~\citep{DRC19} contains more extreme poses, and it consists of two landmark configurations: semi-frontal (68 landmarks) and profile (39 landmarks). The semi-frontal track contains 5,658 training images and 5,335 test images. For the profile track, there are 1,906 and 1,946 images for training and testing, respectively. In this work, we train and test the proposed model on the two tracks separately.

\textbf{300VW}~\citep{SZC15} is a popular benchmark for video-based facial landmark detection, with the same landmark configuration as 300W~\citep{STZ13}. It contains 114 videos, among which 50 are for training and 64 are for testing. The test videos are further divided into three categories based on their level of difficulty: (1) well-lit conditions (31 videos); (2) unconstrained conditions (19 videos); (3) completely unconstrained conditions (14 videos). To validate the robustness of our model, we train and test on the training and test images, respectively, without using any temporal information. To avoid overfitting,  we sample every 5th frame from each training video. Since no face bounding boxes are provided, we use RetinaFace~\citep{DGZ20} to detect bounding boxes.

\textbf{CelebA}~\citep{LLW15} is a large-scale attributes dataset with 202,599 face images. In this work, the images are only used as the unlabeled data in Section~\ref{sec:4.5}.

\subsubsection{Implementation Details}
\label{sec:4.1.2}

\textbf{Supervised Setting}. The face images are cropped according to the provided bounding boxes, then resized to $256 \times 256$. To preserve more context, the bounding boxes of the datasets with 68 landmarks (i.e., 300W, Menpo 2D, and 300VW) are enlarged by 10\%, and the ones with 98 landmarks (i.e., WFLW) are enlarged by 20\%. We use ResNet-18 pretrained on ImageNet as the backbone by default. We also use ResNet-50 and ResNet-101 in some experiments to obtain better results. MobileNets~\citep{SHZ18, HSC19} are also adopted as backbones since they were designed for better efficiency. Adam~\citep{KiB15} is used as the optimizer. The total number of training epochs is 60. The initial learning rate is 0.0001, decayed by 10 at epoch 30 and 50. The batch size is 16. When the network stride varies, the balancing coefficients $\alpha$ and $\beta$ also need to be adjusted accordingly so that the loss values of classification (i.e., $L_S$) and regression (i.e., $L_O$ and $L_N$) are comparable. Concretely, $\alpha$ and $\beta$ are both set to $0.02$, $0.1$, $0.125$, and $0.25$ for stride 16, 32, 64, and 128, respectively. The data augmentation includes translation ($\pm 30$ pixels on the $x$-axis and $y$-axis, $p=0.5$), occlusion (rectangle with maximum $100$ pixels as length, $p=0.5$), horizontal flipping ($p=0.5$), rotation ($\pm 30$ degrees, $p=0.5$), and blurring (Gaussian blur with maximum $5$ radius, $p=0.3$), where $p$ is the probability of execution.

\textbf{Cross-Domain Setting}. This setting consists of three paradigms: generalizable supervised learning (GSL), unsupervised domain adaptation (UDA), and generalizable semi-supervised learning (GSSL). For all paradigms, the 300W training set is used as labeled data and the test sets of 300W, COFW-68, and WFLW-68 are used for evaluation. As can be seen in Figure~\ref{fig:paradigm}, the differences between the three paradigms are mainly in the unlabeled data: (1) GSL conducts evaluation directly after supervised learning, without using any unlabeled data; (2) UDA utilizes the unlabeled data from target domains (in our case, the training sets of COFW-68 and WFLW-68 without labels); and (3) GSSL utilizes unlabeled data that appears in neither the source domain nor the target domain (in our case, it comes from CelebA). As discussed in Section~\ref{sec:3.4}, it is essential to have consistent cropping styles for the cross-domain setting. Specifically, we enlarge the bounding boxes of 300W, COFW-68, and WFLW-68 by 30\%, 30\%, and 20\%, respectively, for consistency. The boxes of WFLW-68 are also adjusted, as stated in Section~\ref{sec:3.4}. The bounding boxes of CelebA are detected by RetinaFace~\citep{DGZ20}, then enlarged by 20\%. The other implementation details are the same as in the supervised setting.

\begin{table}
\centering
\caption{NME (\%) results of PIPNets (w/o NRM) with different network strides on WFLW validation set.}
\begin{tabular}{lccc}
\hline
Method & Net Stride & Heatmap Size & NME (\%)\\
\hline
\hline
PIPNet & 16 & $16\times16$ & 4.03\\
PIPNet  & 32 & $8\times8$ & \textbf{3.94}\\
PIPNet & 64 & $4\times4$ & 3.96\\
PIPNet & 128 & $2\times2$ & 4.10\\
\hline
\end{tabular}
\label{tab:hyperparameter}
\end{table}

\subsubsection{Evaluation Metrics}
\label{sec:4.1.3}

To compare with previous works, we use normalized mean error (NME) to evaluate our models, where the normalization distance is inter-ocular for 300W, 300VW, COFW, COFW-68, WFLW, and WFLW-68. For AFLW, we use image size as the normalization distance, following~\citep{WSC19}. To deal with the large poses and profile faces in Menpo 2D, \citep{DRC19} proposed to use the face diagonal as the normalization distance, which is also adopted in this work for Menpo 2D.

\subsection{Hyperparameters}
\label{sec:4.2}

Two important hyperparameters of PIPNet are the network stride $S$ and the number of neighbors $C$ in the neighbor regression module. In this section, we conduct experiments on WFLW to select appropriate hyperparameters, where 1,500 images from the WFLW training set are randomly selected as our validation set and the rest are used for training (denoted as the sub-training set).

\begin{figure}
  \includegraphics[width=1\linewidth]{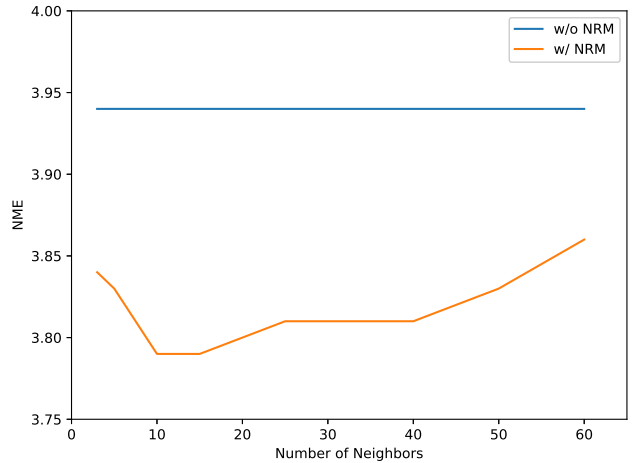}
\caption{NME (\%) results of PIPNets with different number of neighbors in NRM, tested on WFLW validation set. The result of PIPNet w/o NRM is also presented for comparison.}
\label{fig:num_nb}       
\end{figure}

\subsubsection{Network Stride}
\label{sec:4.2.1}

\begin{figure*}
\centering
    \subfigure[\scriptsize{MapNets on NME (\%)}\label{fig:bias_var_1}]{
    \includegraphics[width=0.32\linewidth]{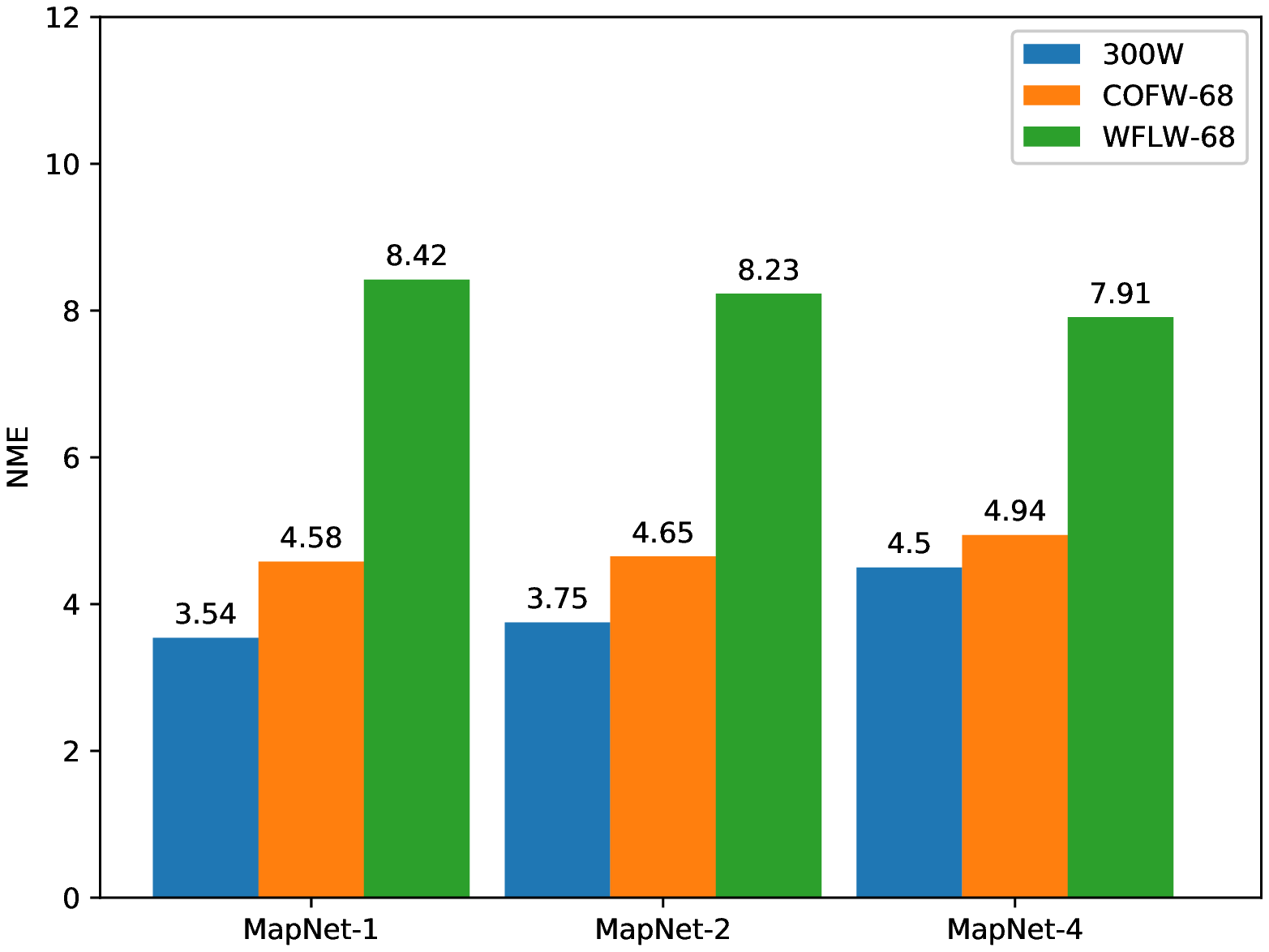}
    }    
    \subfigure[\scriptsize{PIPNets (w/o NRM) on NME (\%)}\label{fig:bias_var_3}]{
    \includegraphics[width=0.32\linewidth]{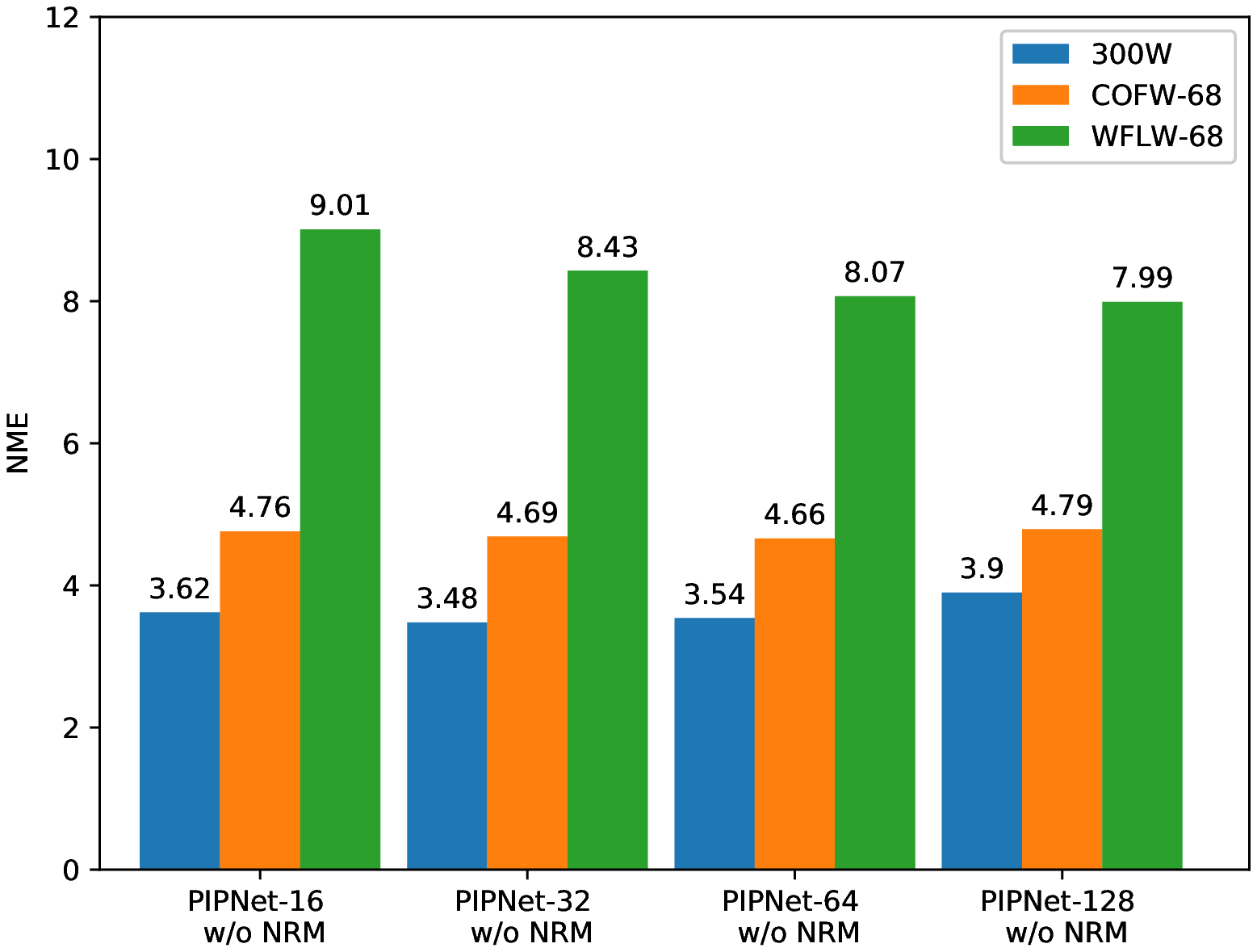}
    }    
    \subfigure[\scriptsize{Comparison with baselines on NME (\%)}\label{fig:bias_var_5}]{
    \includegraphics[width=0.32\linewidth]{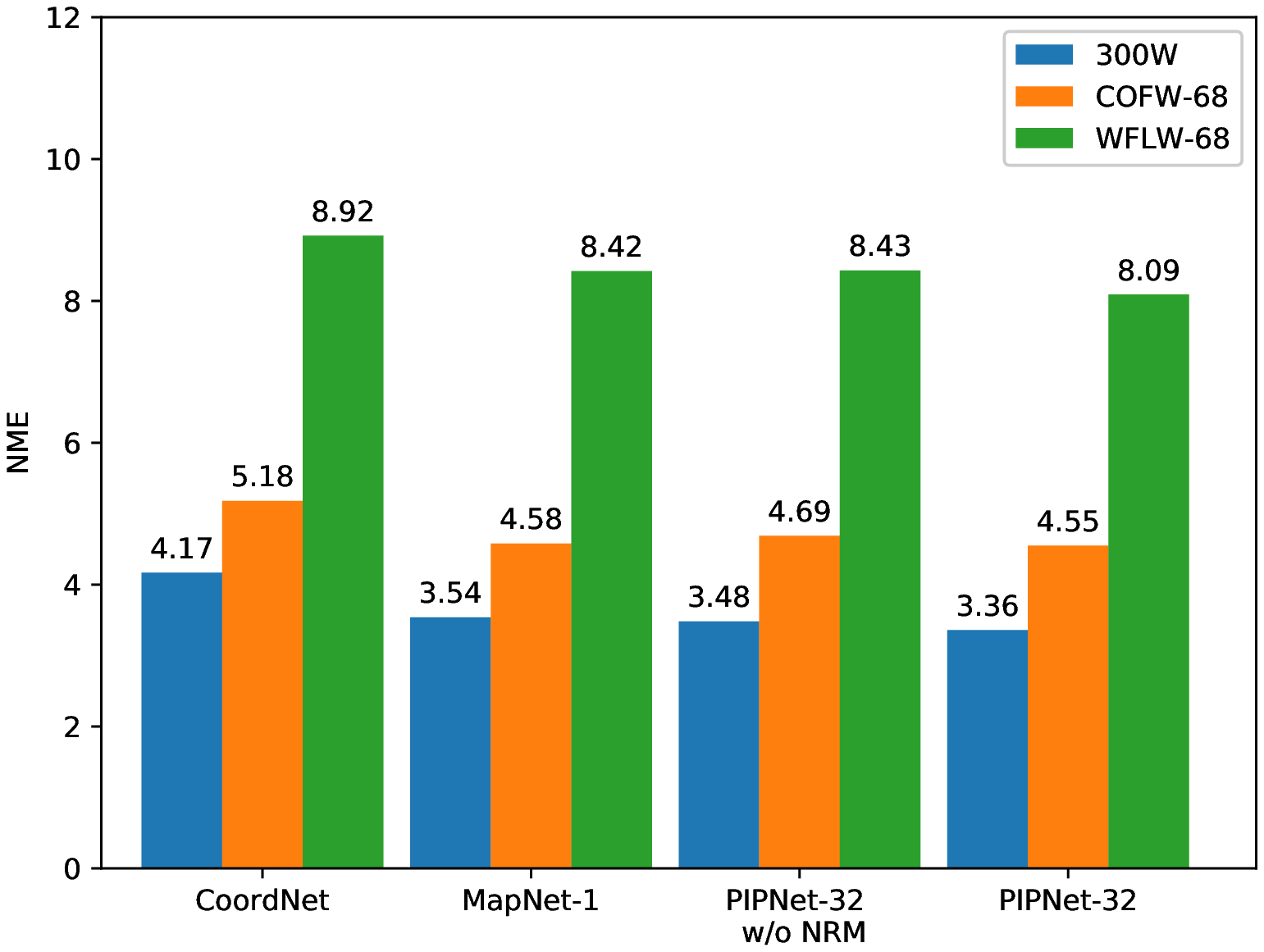}
    } 
    
    \subfigure[\scriptsize{MapNets on Point-Var ($10^{-4}$)}\label{fig:bias_var_2}]{
    \includegraphics[width=0.32\linewidth]{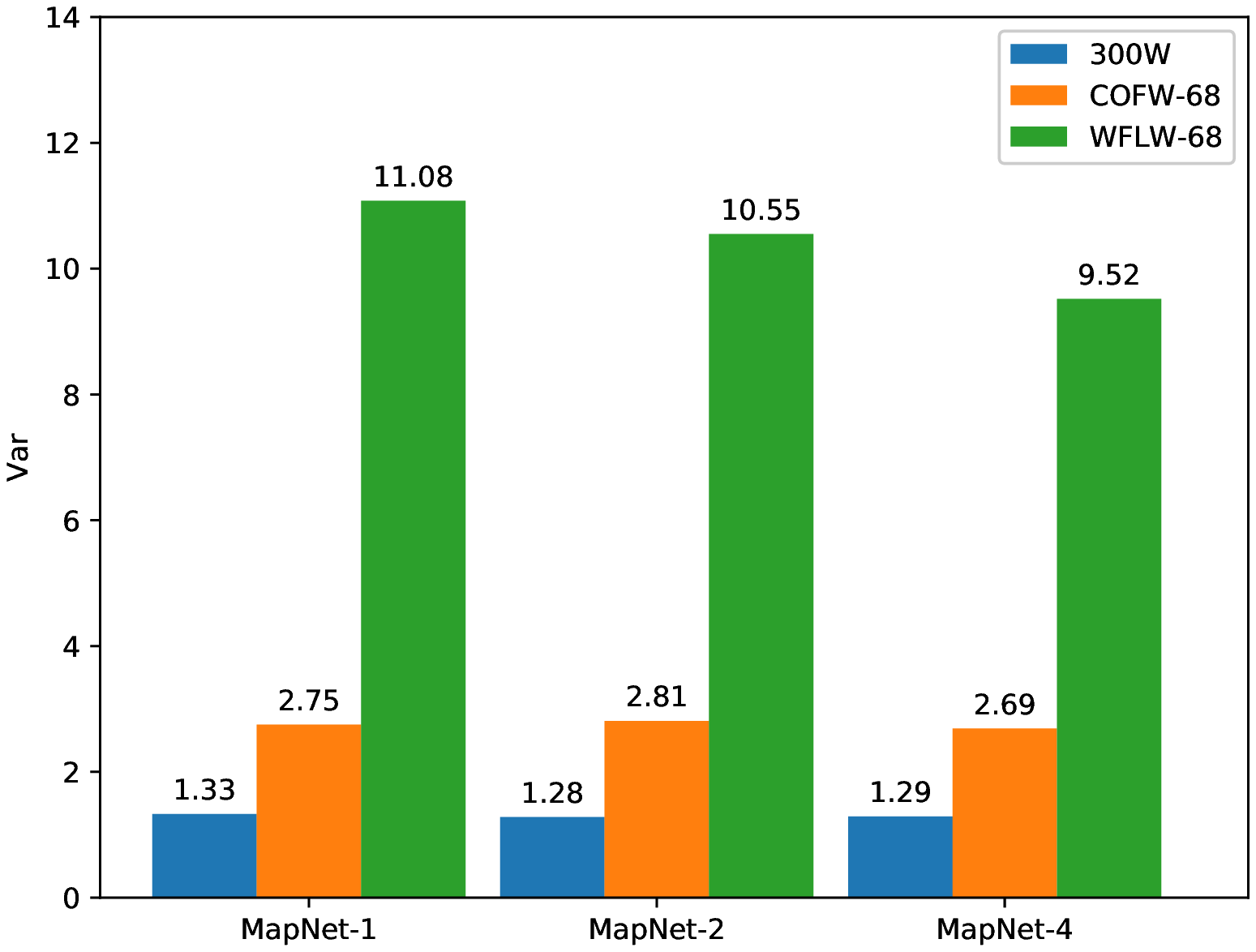}
    }     
    \subfigure[\scriptsize{PIPNets (w/o NRM) on Point-Var ($10^{-4}$)}\label{fig:bias_var_4}]{
    \includegraphics[width=0.32\linewidth]{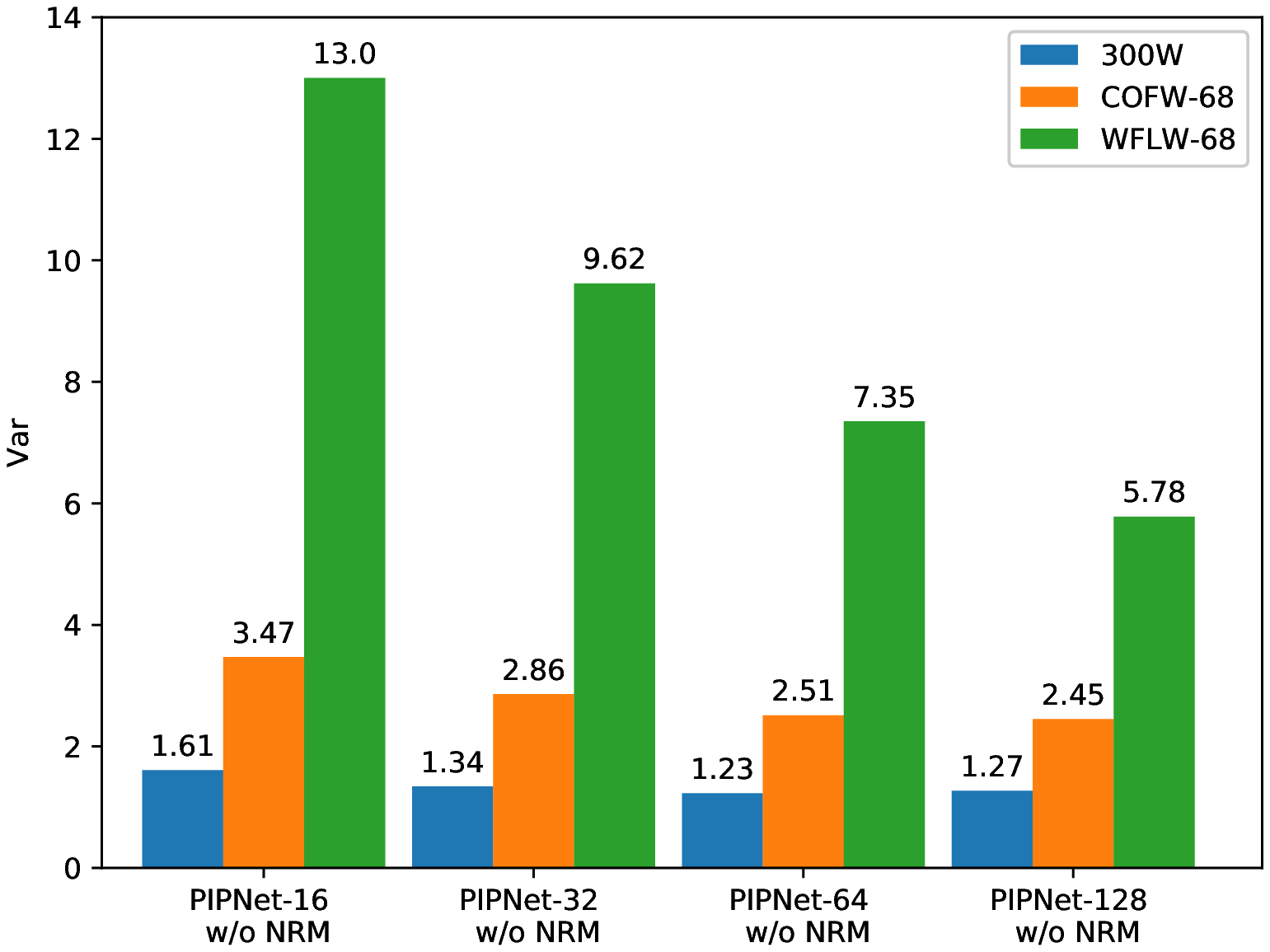}
    }        
    \subfigure[\scriptsize{Comparison with baselines on Point-Var ($10^{-4}$)}\label{fig:bias_var_6}]{
    \includegraphics[width=0.32\linewidth]{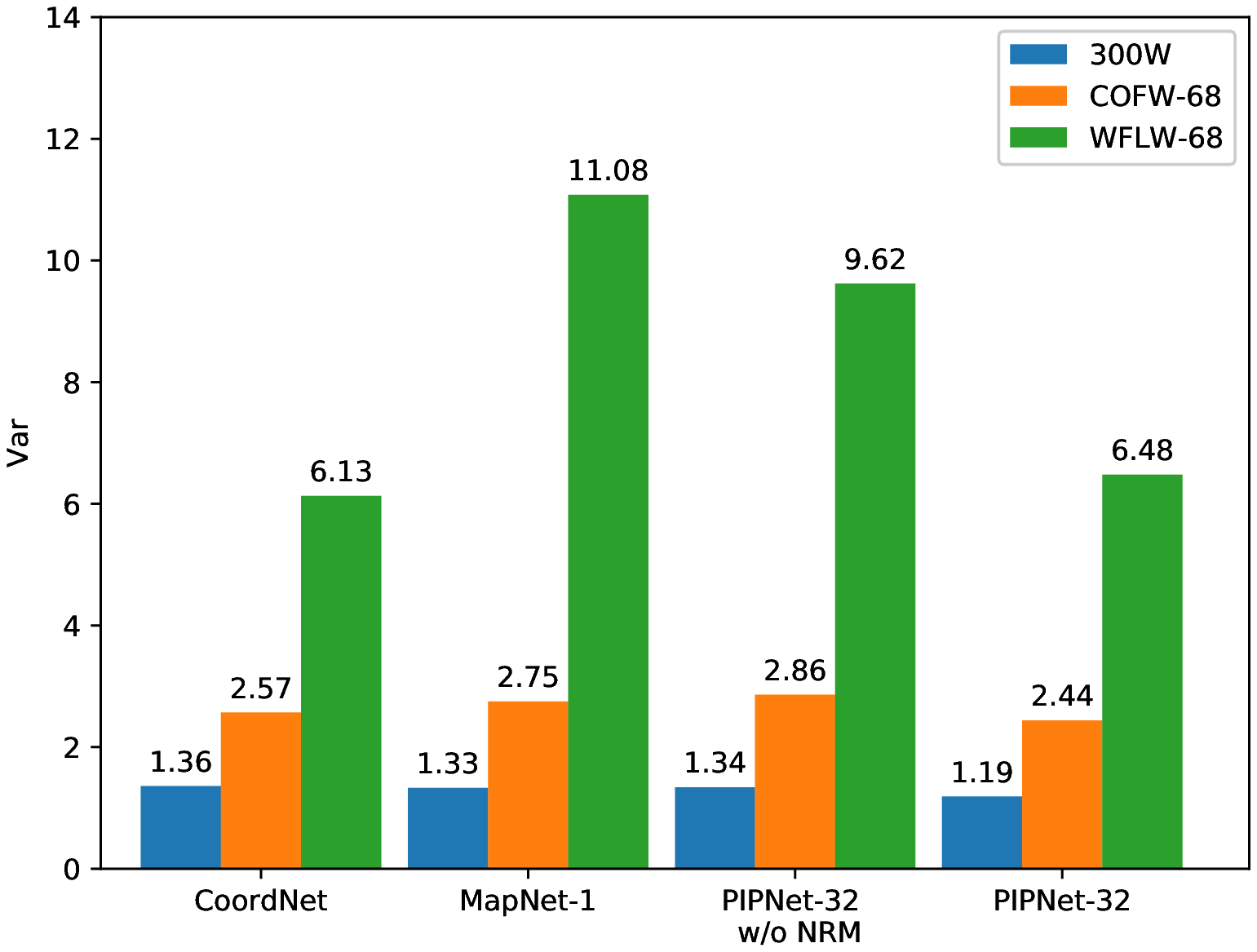}
    }            
    \caption{NME (\%) and Point-Var results of models with various detection heads on 300W, COFW-68, and WFLW-68. (a) NME (\%) of MapNets with different strides. (b) NME (\%) of PIPNets (w/o NRM) with different strides. (c) NME (\%) of three baselines and the proposed model. (d) Point-Var of MapNets with different strides. (e) Point-Var of PIPNets (w/o NRM) with different strides. (f) Point-Var of three baselines and the proposed model.\label{fig:bias_var}}
\end{figure*}

The default stride of ResNet is 32. To get PIPNets with larger strides, we simply add more Conv-BN-ReLU layers, where the convolutional layer is of 512 channels, kernel size $3\times3$, and stride 2. To get smaller strides, the convolutional layer is replaced by a deconvolutional layer with 512 channels, kernel size $4\times4$, and stride 2.  We train PIPNet without the neighbor regression module on the WFLW sub-training set with different network strides. Table~\ref{tab:hyperparameter} shows the results on the WFLW validation set. As can be seen, $S=32$ gives the best result, which will be used for the remaining experiments by default. Intuitively, it provides a trade-off between grid classification and offset regression. As discussed in Section~\ref{sec:3.3}, a larger network stride yields higher accuracy for grid classification, but also results in more errors for offset regression. Therefore, a moderate network stride gives the best performance overall.

\subsubsection{Number of Neighbors}
\label{sec:4.2.2}

To determine an appropriate value of $C$ (number of neighbors in the neighbor regression module), we run PIPNet on the WFLW sub-training set, varying $C$. Figure~\ref{fig:num_nb} shows the NME results on the WFLW validation set. Firstly, we notice that neighbor regression module boosts the performance consistently, which confirms its effectiveness. Because the model achieves the best NME around $C=10$, we use this number for the remaining experiments by default.

\subsection{Model Analysis}
\label{sec:4.3}

In this section, we conduct experiments to analyze the characteristics of the existing detection heads and the proposed one under the generalizable supervised learning paradigm. 

\subsubsection{Baselines}
\label{sec:4.3.1}

To verify the effectiveness of the proposed detection head, we compare it with the existing ones, namely coordinate regression and heatmap regression. We implement CoordNet, which uses coordinate regression as its detection head and ResNet-18 as its backbone. CoordNet consists of three fully connected layers, each of which has $512$, $512$ and $2N$ channels respectively, where $N$ is the number of landmarks. Following~\citep{FKA18}, we use the L1 loss for CoordNet to get better results. For heatmap regression, we choose the model from \citep{XWW18} because it is both effective and lightweight. We implement it as MapNet with ResNet-18 and different network strides. The original model uses a stride of 4 for higher speed. Since we observe an improvement in performance with smaller strides, we also implement MapNet with stride 2 and 1 for a more comprehensive comparison. It is worth noting that MapNet requires Gaussian smoothing on training labels, and the Gaussian radius need to be changed adaptively when the network stride varies so that MapNet can achieve optimal performance. In this work, we use $1$, $2$, and $4$ as the radii for MapNet, with a network stride of $4$, $2$, and $1$, respectively. In contrast, PIPNet does not use Gaussian smoothing, which indicates that PIP regression is easier to train than heatmap regression. The loss function of MapNet is the L2 loss. During the inference stage of MapNet, in addition to the location of the highest response, there is also a quarter offset in the direction from the highest response to the second highest response to make up for the loss of accuracy when the stride is larger than 1~\citep{XWW18}. The rest of the settings for MapNet are the same as PIPNet. In addition to CoordNet and MapNet, we also use PIPNet without the neighbor regression module as a baseline model.

\subsubsection{Bias-Variance Trade-Off}
\label{sec:4.3.2}

Bias and variance are two main sources of prediction error. Due to the trade-off between them, a good model usually minimizes both jointly. According to our observations, coordinate regression gives robust but inaccurate landmark predictions, while heatmap regression is accurate on most samples but sensitive to unusual samples. Therefore, we believe neither is optimally minimized in terms of bias and variance jointly. In order to gain a deeper understanding of this situation, we run experiments on three datasets: 300W, COFW-68, and WFLW-68. All the models are trained on the 300W training set, then directly tested on the test sets of 300W, COFW-68, and WFLW-68. 

Figure~\ref{fig:bias_var_1} gives the NMEs of MapNet with different strides. From the figure, MapNet with stride 1 achieves the lowest NME on 300W and COFW-68, while MapNet with stride 4 performs the best on WFLW-68. A better performance on 300W represents a better capability in fitting data (low bias) because the test data is more similar to the training data. On the other hand, performing better on WFLW-68 means a model has better generalization capability (low variance) because its test images are quite different from the training data. NME contains both bias and variance error, which is not convenient for analyzing the trade-off when comparing different models. Thus, we further compute the variance of the difference between the ground-truth and the predictions for each test image, then average them over a test set to get the Point-Var. Although Point-Var is not exactly the same thing as variance error, it reflects the consistency of predictions, so it can represent variance error to some extent. From the Point-Var results in Figure~\ref{fig:bias_var_2}, we see that the variance on WFLW-68 decreases as the stride increases, which is consistent with Figure~\ref{fig:bias_var_1}. Therefore, we observe that the variance of MapNet decreases as its network stride increases, but the bias also increases significantly.    

\begin{table}
\caption{A summary of the characteristics of various detection heads.}
\begin{tabular}{cccc}
\hline
Detection Head        & Efficient & \begin{tabular}[c]{@{}c@{}}Accurate \\ (Low Bias)\end{tabular} & \begin{tabular}[c]{@{}c@{}}Robust \\ (Low Variance)\end{tabular} \\ \hline
Coordinate  & \cmark       & \xmark & \cmark \\ 
Heatmap     & \xmark        & \cmark  & \xmark \\
PIP         & \cmark       & \cmark   & \xmark \\ 
PIP  + NRM  & \cmark       & \cmark   & \cmark \\ \hline
\end{tabular}
\label{tab:heads_summary}
\end{table}

Figure~\ref{fig:bias_var_3} and \ref{fig:bias_var_4} give the NME and Point-Var results of PIPNets (without NRM) with different strides. Similar to MapNets, the variance decreases as the stride increases. In general, PIPNets (without NRM) have lower bias than MapNets because the bias of PIPNets does not increase significantly as the stride becomes larger. That is to say, PIP regression is a more general framework than heatmap regression when the network stride varies.

To compare between baselines, we choose a representative model from the MapNets and PIPNets (without NRM), respectively, namely MapNet-1 and PIPNet-32 (without NRM). Figures~\ref{fig:bias_var_5} and \ref{fig:bias_var_6} show the NME and Point-Var results of the three baselines and the proposed model on the three test sets. From Figure~\ref{fig:bias_var_5}, we first see that CoordNet performs poorly on all the datasets, which indicates that coordinate regression tends to have high bias. Compared to CoordNet, MapNet and PIPNet (without NRM) have lower bias but higher variance (see Figure~\ref{fig:bias_var_6}). Notably, with the help of the neighbor regression module, PIPNet achieves the lowest NME on all the datasets, which indicates its superiority over the three baselines. From Figure~\ref{fig:bias_var_6}, we see that the variance of PIPNet-32 is comparable to that of CoordNet, which proves the effectiveness of the neighbor regression module on improving model robustness. Table~\ref{tab:heads_summary} summarizes the characteristics of these models. As can be seen, PIP regression + NRM is efficient, accurate, and robust at the same time. Thus, we claim that it possesses the advantages of both coordinate and heatmap regression. 

\subsection{Comparison with State of the Arts}
\label{sec:4.4}

\begin{table*}
\centering
\caption{Comparison with state-of-the-art methods on 300W, COFW, and AFLW. The results are in NME (\%), using inter-ocular distance for normalization. * denotes that the method uses external area data for training. \textcolor{red}{Red} indicates best, and \textcolor{blue}{blue} is for second best.}
\newcolumntype{C}{>{\centering\arraybackslash}X}%
\begin{tabularx}{\textwidth}{lClCCCCC}
\hline\noalign{\smallskip}	
\multirow{2}{*}{Method} & \multirow{2}{*}{Year} & \multirow{2}{*}{Backbone} & \multicolumn{3}{c}{300W} & COFW & AFLW\\ \cline{4-6}
\noalign{\smallskip}
 & & & Full & Com. & Cha. & Full & Full\\
\noalign{\smallskip}	
\hline
\noalign{\smallskip}	
RCN           & 2016 & -  & 5.41   & 4.67   & 8.44   & -    & 5.6 \\ 
DAC-CSR        & 2017 & -  & -      & -      & -      & 6.03 & 2.27    \\ 
TSR                         & 2017 & -  & 4.99   & 4.36    & 7.56    & -    & 2.17 \\
LAB  & 2018 & ResNet-18  & 3.49   & 2.98   & 5.19   & 5.58 & 1.85 \\ 
Wing           & 2018 & ResNet-50 & -   & -   & -   & 5.07    & 1.47 \\ 
SAN           & 2018 & ITN-CPM  & 3.98   & 3.34   & 6.60   & -  & 1.91 \\
RCN+           & 2018 & -  & 4.90   & 4.20   & 7.78   & -    & - \\ 
HG+SA+GHCU     & 2019 & Hourglass  & -      & -      & -      & -    & 1.60 \\ 
TS$^3$ & 2019 & Hourglass+CPM & 3.78 & - & -   & -  & -    \\ 
LaplaceKL      & 2019 & -  & 4.01   & 3.28   & 7.01   & -    & 1.97  \\ 
HG-HSLE        & 2019 & Hourglass  & 3.28   & 2.85   & 5.03   & -    & -    \\ 
ODN            & 2019 & ResNet-18  & 4.17   & 3.56   & 6.67   & -  & 1.63 \\ 
AVS w/ SAN   & 2019 & ITN-CPM  & 3.86   & 3.21   & 6.49   & -    & -    \\
HRNet          & 2019 & HRNetV2-W18  & 3.32   & 2.87   & 5.15   & 3.45 & 1.57 \\ 
AWing       & 2019 & Hourglass & \textcolor{red}{\textbf{3.07}} & 2.72 & \textcolor{red}{\textbf{4.52}} & - & - \\
DeCaFA & 2019 & Cascaded U-net & 3.69 & - & - & - & - \\ 
ADA & 2020 & Hourglass & 3.50 & \textcolor{red}{\textbf{2.41}} & 5.68 & - & - \\ 
LUVLi & 2020 & DU-Net & 3.23 & 2.76 & 5.16 & - & - \\ \hline
LAB$^*$       & 2018 &  ResNet-18 & -   &  -  & -   & $3.92^*$ & \textcolor{red}{\textbf{1.25}}$^*$ \\ 
RCN+$^*$           & 2018 & -  & -   & -   & -   & -    & $1.59^*$ \\ 
DCFE$^*$           & 2018 & -  & $3.24^*$ & $2.76^*$ & $5.22^*$ & - & $2.17^*$ \\
TS$^{3*}$ & 2019 & Hourglass+CPM & $3.49^*$ & - & -   &  - &  -   \\ 
LaplaceKL$^*$      & 2019 & -  & $3.91^*$   & $3.19^*$   & $6.87^*$   & -  & - \\ 
3DDE$^*$          & 2019 & -  & \textcolor{blue}{\textbf{3.13}}$^*$ & \textcolor{blue}{\textbf{2.69}}$^*$ & $4.92^*$ & - & $2.01^*$ \\
DeCaFA$^*$ & 2019 & Cascaded U-net & $3.39^*$ & $2.93^*$ & $5.26^*$ & - & - \\ \hline
PIPNet (ours) & -    & MobileNetV2  & 3.40   & 2.94   & 5.30   & 3.43 & 1.52 \\ 
PIPNet (ours) & -    & MobileNetV3  & 3.36   & 2.94   & 5.07   & 3.40 & 1.52 \\ 
PIPNet (ours) & -    & ResNet-18  & 3.36   & 2.91   & 5.18   & 3.31 & 1.48 \\ 
PIPNet (ours) & -    & ResNet-50  & 3.24   & 2.80   & 5.03   & \textcolor{blue}{\textbf{3.18}} & 1.44 \\
PIPNet (ours) & -    & ResNet-101  & 3.19   & 2.78   & \textcolor{blue}{\textbf{4.89}}   & \textcolor{red}{\textbf{3.08}} & \textcolor{blue}{\textbf{1.42}} \\
\hline
\end{tabularx}
\label{tab:300w-cofw-aflw}
\end{table*}

\begin{table*}
\centering
\caption{Comparison with state-of-the-art methods on WFLW. The NME (\%) results are evaluated on the full set and six subsets: pose set, expression set, illumination set, make-up set, occlusion set, and blur set, using inter-ocular distance for normalization. * denotes that the method uses external area data for training. \textcolor{red}{Red} indicates best, and \textcolor{blue}{blue} is for second best.}
\newcolumntype{C}{>{\centering\arraybackslash}X}%
\begin{tabularx}{\textwidth}{lClCCCCCCCC}
\hline\noalign{\smallskip}	
Method  & Year & Backbone & Pose  & Expr. & Illu. & M.u. & Occ. & Blur & Full\\
\noalign{\smallskip}	
\hline
\noalign{\smallskip}	
LAB     & 2018 & ResNet-18 & 10.24 & 5.51 & 5.23 & 5.15 & 6.79 & 6.32 & 5.27 \\
Wing    & 2018 & ResNet-50 & 8.43  & 5.21 & 4.88 & 5.26 & 6.21 & 5.81 & 4.99 \\
AVS w/ Res-18  & 2019 & ResNet-18 & 9.10 & 5.83 & 4.93 & 5.47 & 6.26 & 5.86 & 5.25 \\ 
AVS w/ LAB  & 2019 & ResNet-18 & 8.21 & 5.14 & 4.51 & 5.00 & 5.76 & 5.43 & 4.76 \\ 
AVS w/ SAN  & 2019 & ITN-CPM & 8.42 & 4.68 & \textcolor{blue}{\textbf{4.24}} & 4.37 & 5.60 & \textcolor{red}{\textbf{4.86}} & 4.39 \\ 
HRNet   & 2019 & HRNetV2-W18 & 7.94  & 4.85 & 4.55 & 4.29 & 5.44 & 5.42 & 4.60 \\
AWing   & 2019 & Hourglass & \textcolor{red}{\textbf{7.38}} & 4.58 & 4.32 & \textcolor{blue}{\textbf{4.27}} & \textcolor{red}{\textbf{5.19}} & \textcolor{blue}{\textbf{4.96}} & \textcolor{blue}{\textbf{4.36}} \\
DeCaFA   & 2019 & Cascaded U-net & - & - & - & - & - & - & 5.01 \\ \
LUVLi   & 2020 & DU-Net & - & - & - & - & - & - & 4.37 \\ \hline
3DDE$^*$    & 2019 & - & $8.62^*$ & $5.21^*$ & $4.65^*$ & $4.60^*$ & $5.77^*$ & $5.41^*$ & $4.68^*$ \\
DeCaFA$^*$   & 2019 & Cascaded U-net & $8.11^*$ & $4.65^*$ & $4.41^*$ & $4.63^*$ & $5.74^*$ & $5.38^*$ & $4.62^*$ \\ \hline
PIPNet (ours) & - & MobileNetV2  & 8.76   & 4.86   & 4.56   & 4.60 & 6.04 & 5.53 & 4.79\\ 
PIPNet (ours) & - & MobileNetV3  & 8.22   & 4.75   & 4.49   & 4.46 & 5.72 & 5.31 & 4.65\\ 
PIPNet (ours) & - & ResNet-18 & 8.02 & 4.73 & 4.39 & 4.38 & 5.66 & 5.25 & 4.57 \\
PIPNet (ours) & - & ResNet-50 & 7.98 & \textcolor{blue}{\textbf{4.54}} & 4.35 & \textcolor{blue}{\textbf{4.27}} & 5.65 & 5.19 & 4.48 \\
PIPNet (ours) & - & ResNet-101 & \textcolor{blue}{\textbf{7.51}} & \textcolor{red}{\textbf{4.44}} & \textcolor{red}{\textbf{4.19}} & \textcolor{red}{\textbf{4.02}} & \textcolor{blue}{\textbf{5.36}} & 5.02 & \textcolor{red}{\textbf{4.31}} \\
\hline
\end{tabularx}
\label{tab:wflw}
\end{table*}

\begin{table}
\centering
\caption{Comparison with the \textbf{three best} teams from the Menpo 2D Challenge on the Menpo 2D benchmark. The NME (\%) results are evaluated using the face diagonal as the normalization distance. \textcolor{red}{Red} indicates best, and \textcolor{blue}{blue} is for second best.}
\newcolumntype{C}{>{\centering\arraybackslash}X}%
\begin{tabularx}{0.48\textwidth}{lClCC}
\hline\noalign{\smallskip}	
Method  & Year & Backbone & Semi-frontal  & Profile \\
\noalign{\smallskip}	
\hline
\noalign{\smallskip}	
J. Yang et al.  & 2017 & Hourglass & \textcolor{red}{\textbf{1.20}} & \textcolor{red}{\textbf{1.72}} \\
Z. He et al.    & 2017 & - & 1.39 & 2.47 \\
W. Wu and S. Yang  & 2017 & VGG-16 & 1.35 & 2.17 \\ \hline
PIPNet (ours) & - & Mob. NetV2  & 1.37 & 2.04 \\ 
PIPNet (ours) & - & Mob. NetV3  & 1.35 & 2.04 \\ 
PIPNet (ours) & - & ResNet-18 & 1.34 & 2.01 \\
PIPNet (ours) & - & ResNet-50 & 1.30 & 1.95 \\
PIPNet (ours) & - & ResNet-101 & \textcolor{blue}{\textbf{1.27}} & \textcolor{blue}{\textbf{1.89}} \\
\hline
\end{tabularx}
\label{tab:menpo2d}
\end{table}

We compare PIPNet with several state-of-the-art methods, including RCN~\citep{HYV16}, DAC-CSR~\citep{FKC17}, TSR~\citep{LSX17}, LAB~\citep{WQY18}, Wing~\citep{FKA18}, SAN~\citep{DYO18}, RCN+~\citep{HMT18}, DCFE~\citep{VBV18}, HG+SA+GHCU~\citep{LZH19}, TS$^3$~\citep{DoY19}, LaplaceKL~\citep{RLZ19}, HG-HSLE~\citep{ZZY19}, ODN~\citep{ZSZ19}, AVS~\citep{QSW19}, HRNet~\citep{WSC19}, 3DDE~\citep{VBV19}, AWing~\citep{WBF19}, DeCaFA~\citep{DBC19}, ADA~\citep{CBG20}, LUVLi~\citep{KMM20}, \citet{YLZ17}, \citet{HZK17}, \citet{WuY17}, TCDCN~\citep{ZLL16b}, CFSS~\citep{ZLL15}, FHR+STA~\citep{TLL19}, TSTN~\citep{LLF17}, and \citet{CBG20}, on six benchmarks, i.e., 300W, COFW, AFLW, WFLW, Menpo 2D, and 300VW. 

Table~\ref{tab:300w-cofw-aflw} shows the NME results on 300W, COFW, and AFLW. From the table, we first observe that PIPNet with ResNet-18 gives slightly better results than the ones with MobileNets. Moreover, PIPNet with ResNet-18 achieves similar or even better results when compared to the best existing methods. Specifically, PIPNet with ResNet-18 achieves 3.31 NME on the full COFW test set, outperforming all the existing methods. On the full AFLW test set, PIPNet with ResNet-18 gets 1.48 NME, beaten only by Wing (1.47 NME) and LAB (1.25 NME). On the full 300W test set, our lightweight model also obtains very competitive result (3.36 NME), and is even better than some methods that use external area data, such as TS$^3$ (3.49 NME, with unlabeled AFLW training data), LaplaceKL (3.91 NME, with 70K unlabeled MegaFace images), and DeCaFA (3.39 NME, with WFLW and CelebA training sets). We notice that most of the state-of-the-art methods use heavyweight backbones like Hourglass, ResNet-50, and HRNetV2-W18. Therefore, we also equip PIPNet with heavier backbones to explore better results. Notably, PIPNet with ResNet-101 achieves state of the art on COFW (3.08 NME), and is significantly better than the best existing model (HRNet, 3.45 NME). On the full AFLW test set, our PIPNet with ResNet-101 is only outperformed by LAB (1.42 vs. 1.25), which uses external boundary data for training. As for 300W, PIPNet with ResNet-101 obtains the second best result on the full test set and challenging set, among the methods that do not use external area data.

Table~\ref{tab:wflw} shows the NME results of the best existing methods and the proposed PIPNets with different backbones on the full WFLW test set and six subsets. As can be seen, our lightweight models (with MobileNetV3 and ResNet-18) already achieve comparable performance to the state of the arts. Again, PIPNet with ResNet-101 obtains the state of the art on the full set (4.31 NME) as well as three subsets. Among the six benchmarks, WFLW is the closest to an uncontrolled environment because it contains more diverse scenes and more in-the-wild images. Consequently, the superior performance on WFLW demonstrates the effectiveness of PIPNet on images in the wild. In Section~\ref{sec:4.6}, we also compare PIPNets with state-of-the-art methods in terms of speed-accuracy trade-off through WFLW results.

To evaluate the compatibility of our model on large poses, we compare PIPNet with the three best teams from the Menpo 2D Challenge~\citep{ZTC17}. Table~\ref{tab:menpo2d} shows the results. Notably, our lightweight models (with MobileNetV3 and ResNet-18) achieve better performance than the second and third best teams~\citep{HZK17,WuY17} on both tracks. Our best model, PIPNet with ResNet-101, is slightly worse than the winner~\citep{YLZ17} of the challenge (1.27 vs. 1.20 on semi-frontal; 1.89 vs. 1.72 on profile). It is worth noting that our models are trained without any specific adaptation to Menpo 2D, while the winner utilized an extra face detector~\citep{CHW16} and facial landmark detector~\citep{BNC16} for first-step transformation since the dataset contains faces with large view angles~\citep{YLZ17}.

\begin{table}
\centering
\caption{Comparison with state-of-the-art methods on 300VW. The NME (\%) results are evaluated with inter-ocular being the normalization distance. \textcolor{red}{Red} indicates best, and \textcolor{blue}{blue} is for second best.}
\newcolumntype{C}{>{\centering\arraybackslash}X}%
\begin{tabularx}{0.48\textwidth}{lClCCC}
\hline\noalign{\smallskip}	
Method  & Year & Backbone & Cat.1  & Cat.2  & Cat.3\\
\noalign{\smallskip}	
\hline
\noalign{\smallskip}	
CFSS  & 2015 & - & 7.68 & 6.42 & 13.7 \\
TCDCN    & 2016 & - & 7.66 & 6.77 & 15.0 \\
TSTN  & 2017 & - & 5.36 & 4.51 & 12.8 \\ 
FHR+STA  & 2019 & Hourglass & 4.42 & 4.18 & 5.98 \\ 
DeCaFA	  & 2019 & Cas. U-net & 3.82 & 3.63 & 6.67 \\
HG+SA+GHCU & 2019 & Hourglass & 3.85 & \textcolor{blue}{\textbf{3.46}} & 7.51 \\
P. Chandran et al.  & 2020 & Hourglass & 4.17 & 3.89 & 7.28 \\ \hline
PIPNet (ours) & - & Mob. NetV2  & 3.10 & 3.60 & 5.60\\ 
PIPNet (ours) & - & Mob. NetV3  & 3.07 & 3.55 & 5.57\\ 
PIPNet (ours) & - & ResNet-18 & \textcolor{blue}{\textbf{3.04}} & 3.51 & \textcolor{blue}{\textbf{5.32}} \\
PIPNet (ours) & - & ResNet-50 & 3.05 & 3.49 & 5.35 \\
PIPNet (ours) & - & ResNet-101 & \textcolor{red}{\textbf{3.03}} & \textcolor{red}{\textbf{3.42}} & \textcolor{red}{\textbf{5.28}}\\
\hline
\end{tabularx}
\label{tab:300vw}
\end{table}

To further validate the robustness of the proposed method, we conduct an evaluation on 300VW. Table~\ref{tab:300vw} shows the results of PIPNets and prior works. First of all, we notice that the performance gaps between PIPNets with different backbones are not significant. This is due to the fact that there is a limited number of identities in the training set, which may lead to overfitting for larger backbones. Despite not using temporal information, PIPNet with ResNet-101 obtains better results than all the existing methods on all three categories. In particular, our best model significantly outperforms prior works on category 1 and 3, which indicates the superiority of our model in accuracy and robustness, respectively. Three sample videos are presented in the supplementary materials to demonstrate the robustness of our models, including the ones trained on 300VW and WFLW with supervised learning and the one trained on CelebA with semi-supervised learning (see Section~\ref{sec:4.5}). 

\begin{table*}
\centering
\caption{Comparison of PIPNets under different training paradigms and strategies on the 300W, COFW-68, and WFLW-68 test sets. The results are in NME (\%), normalized by inter-ocular distance.}
\newcolumntype{C}{>{\centering\arraybackslash}X}%
\begin{tabularx}{\textwidth}{CClCCC}
\hline\noalign{\smallskip}	
\multirow{2}{*}{Paradigm} & \multirow{2}{*}{Method} & \multirow{2}{*}{Unlabeled Training Data} & \multicolumn{3}{ c }{Test Data} \\ \cline{4-6}
\noalign{\smallskip}
& & & 300W & COFW-68 & WFLW-68\\
\noalign{\smallskip}	
\hline
\noalign{\smallskip}	
GSL & - & - & 3.36 & 4.55 & 8.09\\
\hline
\multirow{3}{*}{UDA} & DANN & COFW-68+WFLW-68  & 3.42 (+1.8\%) & 4.55 (-0.0\%) & 8.01 (-1.0\%)\\
& Self-training & COFW-68+WFLW-68  & 3.35 (-0.3\%) & 4.34 (-4.6\%) & 7.45 (-7.9\%)\\
& STC & COFW-68+WFLW-68  & 3.34 (-0.6\%) & 4.28 (-5.9\%) & 7.28 (-\textbf{10.0\%})\\
\hline
\multirow{3}{*}{GSSL} & DANN & CelebA  & 3.43 (+0.3\%) & 4.56 (+0.2\%) & 8.13 (+1.5\%)\\
& Self-training & CelebA  & 3.27 (-2.7\%) & 4.32 (-5.1\%) & 7.77 (-4.0\%)\\
& STC & CelebA  & 3.23 (-\textbf{3.9\%}) & 4.23 (-\textbf{7.0\%}) & 7.53 (-6.8\%)\\
\hline
\end{tabularx}
\label{tab:stc_nme}
\end{table*}

\begin{table}
\scriptsize
\centering
\caption{Comparison with prior works on 300W and COFW-68 test sets. The results are in NME (\%), normalized by inter-ocular distance.}
\newcolumntype{C}{>{\centering\arraybackslash}X}%
\begin{tabularx}{0.48\textwidth}{lClCC}
\hline \noalign{\smallskip}
\multirow{2}{*}{Method} & \multirow{2}{*}{Paradigm} & \multirow{2}{*}{Backbone} & \multicolumn{2}{ c }{Test Data}\\ \cline{4-5}
\noalign{\smallskip}
& & & 300W & COFW-68\\
\noalign{\smallskip}
\hline
\noalign{\smallskip}
LAB & GSL & ResNet-18 & 3.49 & 4.62\\
ODN & GSL & ResNet-18 & 4.17 & 5.30\\
AVS w/ SAN & GSL & ITN-CPM & 3.86 & 4.43\\
\hline
PIPNet (ours) & GSL &ResNet-18 & 3.36 & 4.55\\
PIPNet (ours) & GSSL &ResNet-18 & \textbf{3.23} & \textbf{4.23}\\
\hline
\end{tabularx}
\label{tab:gsl_gssl}
\end{table}

\subsection{Self-Training with Curriculum}
\label{sec:4.5}

We first verify the effectiveness of the proposed self-training with curriculum (STC) strategy by running experiments under the UDA paradigm. Specifically, we use the 300W training set as the only labeled data, and our test data includes the test sets of 300W, COFW-68, and WFLW-68. The training sets of COFW-68 and WFLW-68 are used as unlabeled data. The self-training method without curriculum is used as a baseline for comparison. We also implemented the domain adversarial neural networks (DANN)~\citep{GaL15,GUA16}, a classic UDA method for classification, as another baseline. Furthermore, the results under the GSL paradigm are also presented, where the model is trained on 300W with supervised learning, then evaluated on the test sets without adaptation. From Table~\ref{tab:stc_nme}, we see that PIPNet with ResNet-18 achieves 3.36, 4.55, and 8.09 NME on 300W, COFW-68, and WFLW-68, respectively, under the GSL paradigm. The results indicate large domain gaps between the three datasets, especially WFLW-68. The improvements of DANN against GSL paradigm on COFW-68 (-0.0\%) and WFLW-68 (-1.0\%) are limited, and its performance even degrades on 300W (+1.8\%). This implies that it is difficult to directly apply UDA methods from classification task to facial landmark detection due to the intrinsic discrepancy between the two tasks. When applying the standard self-training method, the domain gaps are considerably reduced, with the NME on COFW-68 and WFLW-68 reduced by 4.6\% and 7.9\%, respectively. With the help of the proposed STC strategy, the NME reduction on COFW-68 and WFLW-68 further becomes 5.9\% and 10.0\%, respectively. As a by-product, the NME of 300W is also slightly improved by 0.6\% due to the increased training data. Thus, the proposed STC is a simple yet effective method that is able to consistently boost the cross-domain performance. STC could also be applied to other vision tasks such as detection. 

\begin{table*}
\centering
\caption{Parameter size, GFLOPs and FPS of existing methods, baselines, and our model.}
\newcolumntype{C}{>{\centering\arraybackslash}X}%
\begin{tabularx}{\textwidth}{lClCCCC}
\hline\noalign{\smallskip}	
Method  & Year & Backbone & \#Param.  & GFLOPs & FPS (CPU) & FPS (GPU)\\
\noalign{\smallskip}	
\hline
\noalign{\smallskip}	
LAB     & 2018 & ResNet-18 & 24.1M+28.3M & 26.7+2.4 & 2.1 & 16.7 \\
Wing    & 2018 & ResNet-50 & 91.0M & 5.5 & 8 & 30 \\ 
AVS w/ LAB  & 2019 & ResNet-18 & 28.3M & 2.4 & 2.1 & 16.7 \\ 
AVS w/ SAN  & 2019 & ITN-CPM & 7.8M+19.4M & 32.7+30.1 & 5.3 & 61 \\ 
HRNet   & 2019 & HRNetV2-W18 & 9.7M & 4.8 & 4.4 & 11.7 \\
AWing   & 2019 & Hourglass & 24.1M & 26.7 & 1.8 & 24.2 \\ 
DeCaFA   & 2019 & Cascaded U-net & 10M & - & - & 32 \\ 
3DDE   & 2019 & - & - & - & - & 12.5 \\
LUVLi   & 2020 & DU-Net & - & - & - & 58.8 \\
\hline
G-RMI & 2017 & ResNet-50 & 23.9M & 25.5 & 0.1 & 6.8 \\
G-RMI & 2017 & ResNet-101 & 42.9M & 45.1 & 0.1 & 5.8 \\
CoordNet & - & ResNet-18 & 28.3M & 2.4 & \textbf{37.5} & \textbf{256} \\
CoordNet & - & ResNet-50 & 91.0M & 5.5 & 14.8 & 115 \\
CoordNet & - & ResNet-101 & 110.0M & 10.4 & 9.4 & 62 \\
MapNet (S=2) & - & ResNet-18 & 28M & 3.0 & 3.5 & 200 \\
MapNet (S=2) & - & ResNet-50 & 52.9M & 6.0 & 2.7 & 82 \\
MapNet (S=2) & - & ResNet-101 & 71.9M & 10.9 & 2.3 & 46 \\ 
\hline
PIPNet (ours) & - & MobileNetV2 & \textbf{4.2M} & 0.5 & 29.5 & 121 \\
PIPNet (ours) & - & MobileNetV3 & 4.5M & \textbf{0.4} & 28.4 & 80 \\
PIPNet (ours) & - & ResNet-18 & 12.0M & 2.4 & 35.7 & 200 \\
PIPNet (ours) & - & ResNet-50 & 26.7M & 5.6 & 13.8 & 99 \\
PIPNet (ours) & - & ResNet-101 & 45.7M & 10.5 & 8.8 & 56 \\
\hline
\end{tabularx}
\label{tab:speed}
\end{table*}

Despite the considerable improvement under UDA, it still may not be ideal for real applications. For example, unlabeled data is not always available for the target domains, or the target domains may even be unknown. Such situations are not uncommon for models running in the wild. Therefore, we aim to go beyond the UDA paradigm and further explore GSSL. To be more specific, the labeled data and test data remain the same as in UDA, but the unlabeled data is changed to CelebA. In this way, the model never sees an image (whether labeled or unlabeled) from the target domain during training. From Table~\ref{tab:stc_nme}, we observe that the results of standard self-training under GSSL are consistently better than those under the GSL paradigm, which indicates the feasibility of GSSL for real applications (i.e., unlabeled data does not necessarily need to be from target domains). In other words, it is feasible to improve the generalization capability of a model using massive amounts of unlabeled data, even if the target domain is unknown. As for DANN, its performance is even worse than the GSL baseline. Again, the STC strategy outperforms standard self-training on all the test sets under GSSL, which confirms its effectiveness. One interesting finding is that the GSSL paradigm obtains better results than UDA on 300W and COFW-68, but worse results on WFLW-68. This is because CelebA is a much larger dataset than COFW-68 and WFLW-68, and its domain is closer to that of 300W and COFW-68. These findings tell us that unlabeled data should not only be collected in large amounts, but also needs to be as diverse as possible under the GSSL paradigm in order to enable models to generalize better on more cross-domain datasets. To demonstrate the superiority of GSSL over GSL, we also compare our model with prior works that conduct direct cross-domain evaluation on COFW-68, including LAB~\citep{WQY18}, ODN~\citep{ZSZ19}, and AVS with SAN~\citep{QSW19}. Table~\ref{tab:gsl_gssl} gives the NME results on the test sets of 300W and COFW-68, where the listed methods are all trained on the labeled 300W training set, with the images from COFW being unavailable during training. As shown in the table, under the same GSL paradigm, PIPNet already obtains the best result on 300W, and is quite competitive on COFW-68 (only inferior to AVS with SAN). Under the GSSL paradigm, PIPNet obtains even better results on 300W (3.23 NME), and outperforms AVS with SAN by 4.5\% (4.23 NME vs. 4.43 NME) on COFW-68, yielding the new best result on the COFW-68 test set. Therefore, GSSL is a promising and scalable paradigm for improving cross-domain generalization ability.

\subsection{Speed}
\label{sec:4.6}

Table~\ref{tab:speed} lists the parameter size, GFLOPs, and speed of the methods in Table~\ref{tab:wflw}. Additionally, we add CoordNets and MapNets equipped with different backbones. The MapNets with stride 2 are used here because they have better speed-accuracy trade-off than the ones with stride 1 and 4. G-RMI~\citep{PZK17} is also implemented as a baseline, where binary cross-entropy and smooth L1 loss are used for classification and regression, and their loss scalars are set to 4 and 1 respectively. For G-RMI, the radius of the disk is set to 15, which gives the best performance. The speeds are averaged over WFLW test set with a batch size of 1, and given in frames per second (FPS). Our code was implemented in PyTorch. The CPU is Intel Xeon E5-2698 v4 @2.20GHz and the GPU is an NVIDIA Tesla V100. From the table, we find that MobileNets are not as efficient as ResNet-18, especially on GPU, although they have smaller GFLOPs. We believe this is related to their implementation in PyTorch, which is not fully optimized. Therefore, we use ResNets as the backbones of PIPNet for the comparison of speed-accuracy trade-off. Figures~\ref{fig:speed_cpu} and \ref{fig:speed_gpu} show the speed-accuracy trade-off comparison of the existing methods, baselines, and PIPNet on CPU and GPU, respectively. The NME results are obtained on the WFLW test set with an image size of 256$\times$256. The existing methods with * were tested by us under the same environment as our models. As we can see, PIPNet achieves the best speed-accuracy trade-off on both CPU and GPU, thanks to the lightweight detection head. G-RMI obtains comparable NMEs to PIPNet, but its speed is about 100$\times$ and 10$\times$ slower on CPU and GPU respectively, due to the heavy computations on high-resolution feature maps. Similarly, MapNet is much slower than PIPNet on CPU due to its heavy detection head, although the NMEs are satisfactory. Interestingly, MapNet becomes much faster on GPU, and we believe this is because the deconvolutional layers are highly optimized in PyTorch with GPU. In contrast, CoordNet is faster than PIPNet, but its accuracy is significantly worse due to the biased predictions. It is worth noting that PIPNet with ResNet-18 is the only model that runs in real-time (35.7 FPS) on CPU while still achieving competitive result to state-of-the-art methods.

\section{Conclusion}
\label{sec:5}

In this work, we propose a novel facial landmark detection framework named PIPNet. PIPNet consists of three new modules, namely PIP regression, neighbor regression, and self-training with curriculum. PIP regression is a lightweight detection head based on heatmap regression. To be more specific, it only predicts low-resolution score heatmaps, where each heatmap pixel further predicts offsets within itself to yield accurate predictions. By eliminating the repeated upsampling layers, the proposed detection head saves considerable computational cost, especially on lightweight computing devices. The neighbor regression module is another lightweight module that aims to improve model robustness by fusing the predictions from neighboring landmarks. Self-training with curriculum is a new strategy that can utilize unlabeled data across domains. By gradually increasing the difficulty of the tasks for pseudo-labeled data, self-training with curriculum introduces less errors from the estimated pseudo-labels, enabling the model to generalize better on cross-domain datasets. In summary, extensive experiments show that PIPNet is an efficient, accurate, and robust facial landmark detector that can run in the wild on lightweight devices.


\bibliographystyle{spbasic}      
\bibliography{mybib}   

\end{document}